\documentclass[10pt,twocolumn]{article}
\usepackage{hyperref}
\hypersetup{
colorlinks=true,
urlcolor=blue,
citecolor=blue}
\usepackage[all]{xy,xypic}
\usepackage{amsfonts,amssymb,amsmath,amsgen,amsopn,amsbsy,theorem,graphicx,epsfig}
\usepackage{eufrak,amscd,bezier,latexsym,mathrsfs,eurosym,enumerate}
\usepackage[utf8]{inputenc}
\usepackage[english]{babel}
\usepackage{cleveref,multirow}
\usepackage[dvipsnames]{xcolor}
\usepackage[pagewise]{lineno}
\usepackage{subcaption}

\usepackage{cuted}
\providecommand{\keywords}[1]
{
  \small	
  \textbf{\textit{Keywords---}} #1
}

\title{Anomaly Localization in Model Gradients \\ Under Backdoor Attacks Against Federated Learning}
\author{Zeki BILGIN \\  
Arcelik Research, Istanbul, Turkey \\
zeki.bilgin@arcelik.com} 

\date{} 
\begin{document}

\maketitle

\begin{abstract}

Inserting a backdoor into the joint model in federated learning (FL) is a recent threat raising concerns. Existing studies mostly focus on developing effective countermeasures against this threat, assuming that backdoored local models, if any, somehow reveal themselves by anomalies in their gradients. However, this assumption needs to be elaborated by identifying specifically  which gradients are more likely to indicate an anomaly to what extent under which conditions. This is an important issue given that neural network models usually have huge parametric space and consist of a large number of weights. In this study, we make a deep gradient-level analysis on the expected variations in model gradients under several  backdoor attack scenarios against FL. Our main novel finding is that backdoor-induced anomalies in local model updates (weights or gradients) appear in the final layer bias weights of the malicious local models. We support and validate our findings by both theoretical and experimental analysis in various FL settings. We also investigate the impact of the number of malicious clients, learning rate, and malicious data rate on the observed anomaly. Our implementation is publicly available\footnote{\url{ https://github.com/ArcelikAcikKaynak/Federated_Learning.git}}.    


\end{abstract}

\section{Introduction}

Federated Learning (FL) is a technique of collaborative learning among many participants, which allows to develop a joint machine learning (ML) model trained on decentralized data \cite{mcmahan2017communication}. It  does not require participants to expose or exchange their local data while generating a joint model, and thus preserves data privacy. 

In an FL setting, there is usually a central unit or server/aggregator connected to multiple users or clients. The server manages the learning procedure in an iterative manner as follows: At the beginning, it constructs a preliminary model and sends the whole model or just its weights to the clients. Upon receiving such a model from the server, the clients update this model by training it with their local data and send back to the server.  The server calculates the joint (global) model by taking the average of the model weights received from the clients. This is one round. Then the server initiates another round by sending the most recent joint model to clients. Upon receiving a more recent  joint model, the clients repeat  the same procedure by training the newly received joint model by their local data and send back to the server. The procedure repeats like this until being stopped depending on some criteria such as joint model accuracy, loss, or convergence rate.

Because FL involves many participants in collaborative model training, it becomes vulnerable to certain threats such as model poisoning or backdoor attacks \cite{wang2020attack,xie2019dba,lyu2020threats}. In a traditional poisoning attack, it is usually aimed to degrade the reliability of a model by manipulating training data, whereas, in a backdoor attack, the objective becomes to insert a hidden trigger into the model that gives an advantage to the adversary when an input with certain features is given to the model \cite{truong2020systematic}. As an example for the backdoor attack, it is demonstrated in  \cite{bagdasaryan2020backdoor} that a malicious participant can manage the joint image classification model to misclassify all cars with a racing stripe as ``bird'' on CIFAR-10 dataset \cite{CIFAR10krizhevsky2009learning} while preserving the model accuracy on other inputs. In this study, we give our attention to backdoor attacks. 

Despite the interest in FL  has been increasing every day to ensure data privacy and fulfill legal obligations in this regard such as the EU General Data Protection Regulation (GDPR) \cite{voigt2017eu}, security threats, some of which are mentioned above, highlight the need for secure FL \cite{andreina2020baffle,sun2019can,nguyen2021flguard,Bilgin2021}. In this sense, although there are plenty of studies presenting secure FL solutions from different perspectives, we believe that the anomalies at the level of model weights caused by the backdoor attacks  have not been studied sufficiently in terms of localization.  With the objective of contributing to fill this gap, in this study, we examine likely anomalies in model gradients that may occur during a backdoor attack initiated by one or multiple malicious client(s) in FL. Our main contribution is to reveal by theoretical and experimental evidence that  exactly which weights in local models can be expected to indicate abnormal deviations caused by an attempted backdoor attack  during training rounds under various FL settings.

\section{Related Work}
\label{sec:relatedwork}

The study \cite{gao2020backdoor} provides a comprehensive review of backdoor attacks within the scope of general deep learning settings. The authors of \cite{ji2017backdoor} draw attention to the threat of backdoor attacks targeting machine learning systems and provide explanations for the success of these attacks. Unlike the approach of creating a backdoor attack by just mislabeling the training data with certain features, the authors of \cite{saha2020hidden} demonstrate a form of backdoor attack that is triggered by adding a small patch into the images to be tested in deep learning models. The work \cite{truong2020systematic} presents a systematic assessment of the effectiveness of backdoor attacks on image classifiers under different experimental conditions such as trigger patterns, poisoning strategies, model architecture, along with potential defensive regularization techniques. In \cite{wang2019neural}, a backdoor detection method is presented for deep neural networks with several mitigation techniques such as input filters, neuron pruning, and unlearning. Unlike the backdoor attacks targeting image classifiers, a backdoor attack to graph neural networks is demonstrated in  \cite{zhang2020backdoor}   for graph classification. 

The backdoor attacks briefly mentioned above are not specifically designed for FL.
In the scope of FL, the study \cite{nguyen2021flguard}  presents a dynamic clustering approach to identify backdoored local models by calculating the pairwise Cosine distances measuring the angular differences between all model updates. In this approach, equal importance is given to all gradients while calculating the differences between local model updates, however, focusing on a certain group of  gradients may be more effective and consistent for anomaly detection as we show in our study. In  
\cite{sun2019can}, the authors  apply  norm clipping and “weak” differential privacy over local model weights to  mitigate the backdoor attacks in FL.  
The authors of \cite{wu2020mitigating} perform neuron pruning to mitigate backdoor attacks by simplifying the model architecture as eliminating redundant neurons, and thus they attempt to remove possible backdoor impacts by modifying model weights and network architecture without required to detect malicious (backdoored) local models.
From a different perspective, the work \cite{andreina2020baffle} brings a solution against backdoor attacks in FL that consists of a round-based feedback loop engaging clients in the performance evaluation of the current joint model with respect to the earlier version. Depending on such a comparative performance evaluation, the server decides whether to accept or discard the updated joint model, with the objective of invalidating possible poisonous updates. However, this approach may also prevent benefiting from useful updates while  eliminating possibly harmful ones.          In \cite{Bilgin2021}, the authors develop a secure aggregation protocol suitable for use against backdoor attacks in FL  based on multi-party computation such that the protocol allows the server to detect model weights larger than a predetermined
threshold value in a privacy-preserving manner without learning any information about
the weights.

\section{Threat Model}

\begin{figure}[h!]
\begin{center}
\includegraphics[width=\linewidth]{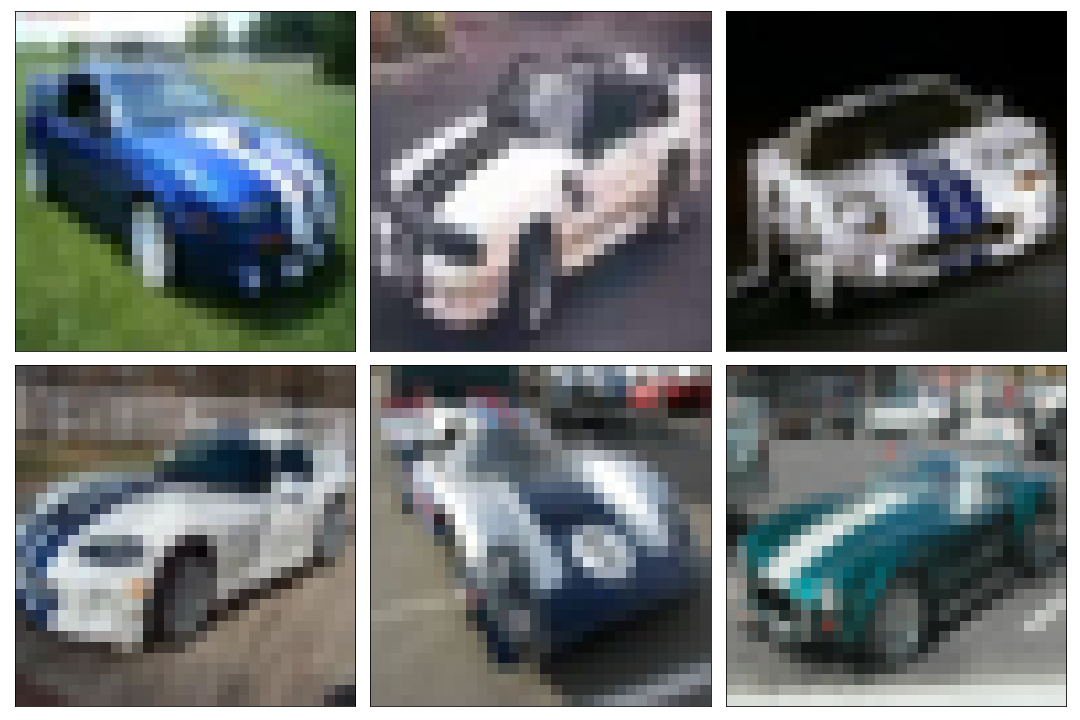}
\caption{Sample CIFAR10 images to be used for the backdoor insertion into the joint model.}
\label{fig:backdoordata}
\end{center}
\end{figure}

In federated learning, there is a server node and multiple connected clients. The clients collaboratively and iteratively build a joint machine learning model that is maintained at the server. Building a joint model takes place in iterative training rounds. In one training round, the server distributes the joint model to the clients, and the clients update (train) the received joint model by their local data, and then send the updated model back to the server. The server generates the joint model by taking average of the local models received at the end of a round. In such a FL setting, a common way  to realize a backdoor attack is to use purposefully mislabeled training data such that a portion of the training data with certain features is mislabeled to targeted class or value in the malicious clients. Likewise, in this work, to simulate a malicious client, we intentionally mislabel  the images of cars with stripe as bird, as similarly demonstrated in \cite{bagdasaryan2020backdoor}. In such a scenario, it is expected that the joint model maintains to recognize regular car images accurately while misrecognizing the cars with stripes as bird.  Figure \ref{fig:backdoordata} shows a few samples of such images to be used for backdoor attack  from CIFAR10 dataset.   

\section{Mathematical Analysis}
\label{sec:theory}

\begin{figure*}[h!]
\begin{center}
\includegraphics[width=0.7\linewidth]{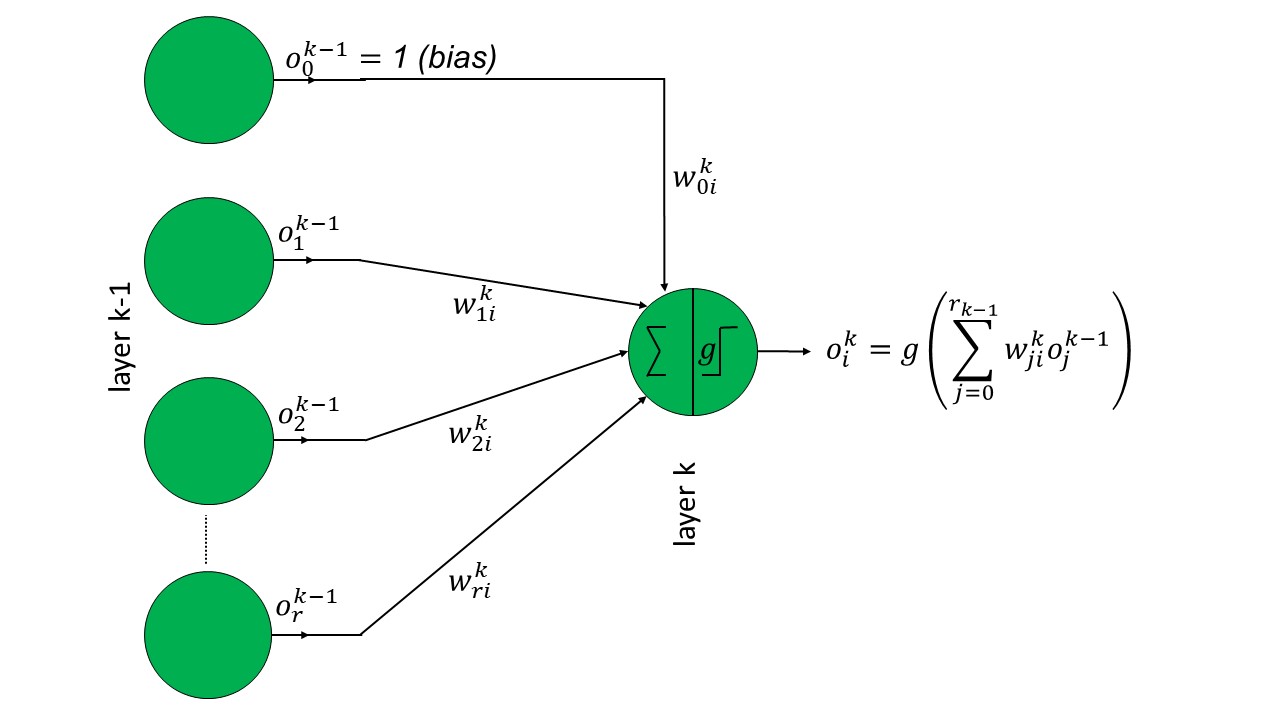}
\caption{Perceptron}
\label{fig:perceptron}
\end{center}
\end{figure*}

In a neural network using backpropagation, model weights are updated layer by layer in backward direction starting from the output layer to the input layer as taking into account the error between actual output and desired output. While doing this, the share of each weight in the error   is calculated across the whole network, and weight updates are made so that this error is expected to be reduced. Below, we present how this calculation is done based on well-established mathematical foundations of backpropagation \cite{Brilliant,hecht1992theory} by following the same notation given in \cite{Brilliant}. Table \ref{tab:notation}
includes explanation of the notation and Figure \ref{fig:perceptron} shows a perceptron which is a building block of neural networks.

\begin{table}[h!]
\caption{The notation  used in our mathematical analysis }
\centering
\begin{tabular}{ l|p{6cm} } 
 \hline
 Symbol & Meaning   \\ 
 \hline
 $w_{ij}^k$ & The weight between nodes $i$ and $j$ in layers $l_{k-1}$ and $l_{k}$ respectively  \\ 
 $b_j^k$ & The bias for node $j$ in layer $l_k$  \\ 
 $a_j^k$ & Product sum plus bias for perceptron $j$ in layer $l_k$ \\
 $o_j^k$ & Output for node $j$ in layer $l_k$ \\
 $r_k$ & Number of nodes in layer $l_k$\\
 $f^{\prime}(x)$ & The derivative of a function $f(x)$\\
 \hline
\end{tabular}
\label{tab:notation}
\end{table}

Let $X$ be training set $X = \{(x_1, y_1), (x_2,y_2),.....,(x_N, y_N)\}$, where $(x_i,y_i)$ is a pair of input and corresponding output for $i = 1,2,...,N$. What a neural network does over such a dataset is the approximation of a mapping from input data to output values \cite{hecht1992theory}. The error in this approximation is measured as a function of the network's weights ($W$) depending on the training dataset by the deviation (e.g. the mean squared error) between  the desired output $y_i$  and the actual output $\hat{y_i}$ for a set of input-output pairs $\big(x_i, y_i\big) \in X$  as formulated in Equation \ref{eq:error}. 

\begin{equation}
 E(X, W) = \frac{1}{2N} \sum_{i=1}^N \left(\hat{y_i} - y_i\right)^2
\label{eq:error}
\end{equation}

To reduce the error by updating the weights, it is necessary to find in which direction  each weight should be changed (i.e. decrease or increase). This is achieved by \textit{gradient descent} that finds the gradient of a function at a particular value (e.g. $\frac{\partial E}{\partial w_{ij}^k}$) and then updates that value by moving in the direction of the negative of the gradient   \cite{Brilliant_gradient}. This is done for each weight $w_{ij}^k$ as this derivative indicates the direction of change of the error function with respect to  the change in the corresponding weight, and thus the weights are changed by an amount that is proportional to the size of the error. Notice that this can be done separately for each pair of input-output data over individual error terms or collectively for a certain number of input-output pairs (i.e. as batches) over cumulative error as formulated in Equation \ref{eq:cumlativederive}.

\begin{equation}
    \frac{\partial E(X,W)}{\partial w_{ij}^k} =  \frac{1}{N} \sum_{d=1}^N \frac{\partial E_d}{\partial w_{ij}^k} 
\label{eq:cumlativederive}
\end{equation}

Now, let us take a closer look at the derivative term in Equation \ref{eq:cumlativederive} by benefiting from the inner mechanism of a perceptron, which is the building block  of neural networks. Figure \ref{fig:perceptron}  shows a perceptron, where the product sum of a node's weights with the previous layer's outputs is called the activation of this node and calculated as formulated in Equation \ref{eq:productsum}. By using the activation, the  derivative term in Equation \ref{eq:cumlativederive} can be expressed as the product of two different derivative terms thanks to the chain rule as shown in Equation \ref{eq:chainrule}.

\begin{equation}
    a_i^k = \sum_{j = 0}^{r_{k-1}} w_{ji}^k o_j^{k-1}
\label{eq:productsum}
\end{equation}

\begin{equation}
\frac{\partial E}{\partial w_{ij}^k} =  \frac{\partial E}{\partial a_{j}^k} \frac{\partial a_{j}^k}{\partial w_{ij}^k} 
\label{eq:chainrule}
\end{equation}

where $a_j^k$ is the activation  of node $j$ in layer $k$ before it is passed to the nonlinear activation function ($g$) (e.g. sigmoid or relu) to generate the output. The first term in Equation \ref{eq:chainrule} is usually called the error and denoted as given in Equation \ref{eq:theerror}.

\begin{equation}
\delta_j^k \equiv \frac{\partial E}{\partial a_{j}^k}
\label{eq:theerror}
\end{equation}

On the other hand, when we rewrite the second term in Equation \ref{eq:chainrule} by combining with Equation \ref{eq:productsum} as given in Equation \ref{eq:theerror2}, we obtain the output $o_i^{k-1}$ of node $i$ in layer $k-1$. 

\begin{equation}
\frac{\partial a_{j}^k}{\partial w_{ij}^k} =   \frac{\partial }{\partial w_{ij}^k} \left(\sum_{l=0}^{r_{k-1}} w_{lj}^k o_l^{k-1} \right) = o_i^{k-1}
\label{eq:theerror2}
\end{equation}

Thus, the partial derivative of the error function $E$ with respect to a weight $w_{ij}^k$ is obtained as given in Equation \ref{eq:partialderivative1}, which is the product of the error term $\delta_j^k$ at node $j$ in layer $k$, and the output $o_i^{k-1}$ of node $i$ in layer k-1. This is reasonable because  the weight $w_{ij}^k$ is located between the output of node $i$ in layer $k-1$ and the input of node $j$ in layer $k$.  

\begin{equation}
\frac{\partial E}{\partial w_{ij}^k} = \delta_j^k o_i^{k-1} 
\label{eq:partialderivative1}
\end{equation}

\textbf{The Output Layer}

In backpropagation, computation of the error terms proceeds backward from the output layer down to the input layer. Without loss of generality, let us continue our mathematical analysis by assuming that the error function is the mean squared error, the activation function is sigmoid, and there is only one output node. Thus we can define the error function $E$ as in Equation \ref{eq:outputlayererror}. 

\begin{equation}
E =  \frac{1}{2} (\hat{y} - y)^2 = \frac{1}{2}(g(a_1^m) - y)^2 
\label{eq:outputlayererror}
\end{equation}

where $m$ is the final layer and  $g(x)$ is the activation function for the output layer.
 
We obtain Equation \ref{eq:outputlayererror2} from Equatıon \ref{eq:theerror} as applying the partial derivative.

\begin{equation}
\delta_1^m = (g (a_1^m) - y) g^{'} (a_1^m) = (\hat{y} - y) g^{'} (a_1^m)
\label{eq:outputlayererror2}
\end{equation}

Combining all of them, the partial derivative of the error function $E$ with respect to a weight in the final layer $w_{i1}^m$ becomes as given in Equation \ref{eq:outputlayererror3}.

\begin{equation}
\frac{\partial E}{\partial w_{i1}^m} = \delta_1^m o_i^{m-1} = (\hat{y} - y)  g^{'} (a_1^m) o_i^{m-1}
\label{eq:outputlayererror3}
\end{equation}

\textbf{The Hidden Layers}

The computation of the error terms in hidden layers is a bit different than the output layer as it also depends on the error terms in the next forward layer. Now consider the following equation for the error term $\delta_j^k$ in layer $1 \le k < m$:

\begin{equation}
\delta_j^k = \frac{\partial E}{\partial a_{j}^k} = \sum_{l=1}^{r^{k+1}} \frac{\partial E}{\partial a_{l}^{k+1}} \frac{\partial a_l^{k+1}}{\partial a_{j}^k}
\label{eq:hiddenerror1}
\end{equation}

where $r^{k+1}$ is the number of nodes in the next layer of ${k+1}$ and  $l$ ranges from 1 to $r^{k+1}$ skipping the value 0 because the bias input $o_0^k$  corresponding to $w_{0j}^{k+1}$ is fixed and its value is not dependent on the outputs of previous layers. Given that first partial derivative term in the product sum of Equation \ref{eq:hiddenerror1} is equivalent to  the error term $\delta_l^{k+1}$, Equation   \ref{eq:hiddenerror1} can also be written as in Equation \ref{eq:hiddenerror2}:

\begin{equation}
\delta_j^k = \sum_{l=1}^{r^{k+1}} \delta_l^{k+1} \frac{\partial a_l^{k+1}}{\partial a_{j}^k}
\label{eq:hiddenerror2}
\end{equation}

The second term in Equation \ref{eq:hiddenerror2} can be rewritten as in Equation \ref{eq:hiddenerror3} by using the definition of $a_l^{k+1}$.  

\begin{equation}
\frac{\partial a_l^{k+1}}{\partial a_{j}^k} = w_{jl}^{k+1} g^{'}(a_j^k)
\label{eq:hiddenerror3}
\end{equation}

where $g(x)$ is the activation function for the hidden layers. 

Putting this into the above equation gives a final equation for the error term $\delta_j^k$   in the hidden layers as in Equation \ref{eq:hiddenerror4} \cite{Brilliant}:

\begin{equation}
\delta_j^k = \sum_{l=1}^{r_{k+1}} \delta_j^{k+1} w_{jl}^{k+1} g^{'}(a_j^k)
\label{eq:hiddenerror4}
\end{equation}

\textbf{Overall}

Among the above formulas, we are more concerned with the gradient calculations because model weights are updated in proportion to the size of the associated gradient as given in Equations \ref{eq:overall1}-\ref{eq:overall2}.

\begin{equation}
w_{new} = w_{old} - \alpha \frac{\partial E(X,W)}{\partial w_{old}} 
\label{eq:overall1}
\end{equation}

\begin{equation}
b_{new} = b_{old} - \alpha \frac{\partial E(X,W)}{\partial b_{old}} 
\label{eq:overall2}
\end{equation}

where $w_{old},w_{new},  b_{old}$ and $b_{new}$ are old and updated weight and bias values respectively, and $\alpha$ is the learning rate. Since $\alpha$ is the same for all data and local models in an iteration of  gradient descent, any abnormal deviation in weight or bias updates is expected to be due to gradient.
Equation  \ref{eq:finalall} includes the gradient formulas for bias weights and regular weights at both output layer and hidden layers. Notice that the output of previous layer is 1 for bias weights as reflected in Equation  \ref{eq:finalall}. 

It is seen from Equation  \ref{eq:finalall} that the difference between the actual output of a model and the desired output (i.e.$(\hat{y} - y)$) is a factor that directly affects the value of a gradient for the output layer. Moreover, this is more apparent for bias weights with respect to regular weights at the output layer because the bias gradient depends on only two factors which are the output error ($\hat{y} - y$) and the derivative of activation function ($g^{'} (a_1^m)$), whereas there is an additional third factor of the output of previous layer ($o_i^{m-1}$) for regular weights. 
Therefore, given that a typical backdoor attack is carried out by mislabeling training data which results in higher error at the output, we can expect to see higher gradient values especially for bias gradients at the output layer.

\begin{strip}
\begin{equation}
\frac{\partial E(X,W)}{\partial w_{ij}^k} =  
\begin{cases}
 \frac{1}{N} \sum_{d=1}^N [(\hat{y} - y) g^{'} (a_1^m)]_d  &\text{for output layer bias weight}\\
 \frac{1}{N} \sum_{d=1}^N [(\hat{y} - y) g^{'} (a_1^m) o_i^{m-1}]_d &\text{for output layer regular weights}\\
 \frac{1}{N} \sum_{d=1}^N [g^{'}(a_j^k) \sum_{l=1}^{r_{k+1}} w_{jl}^{k+1} \delta_l^{k+1}]_d &\text{for hidden layer bias weight } \\
 \frac{1}{N} \sum_{d=1}^N [g^{'}(a_j^k) o_i^{k-1} \sum_{l=1}^{r_{k+1}} w_{jl}^{k+1} \delta_l^{k+1}]_d &\text{for hidden layer regular weights } 
\end{cases}
\label{eq:finalall}
\end{equation}
\end{strip}

\section{Experimental Analysis}

\begin{figure*}[h]
\begin{center}
\includegraphics[width=0.6\linewidth]{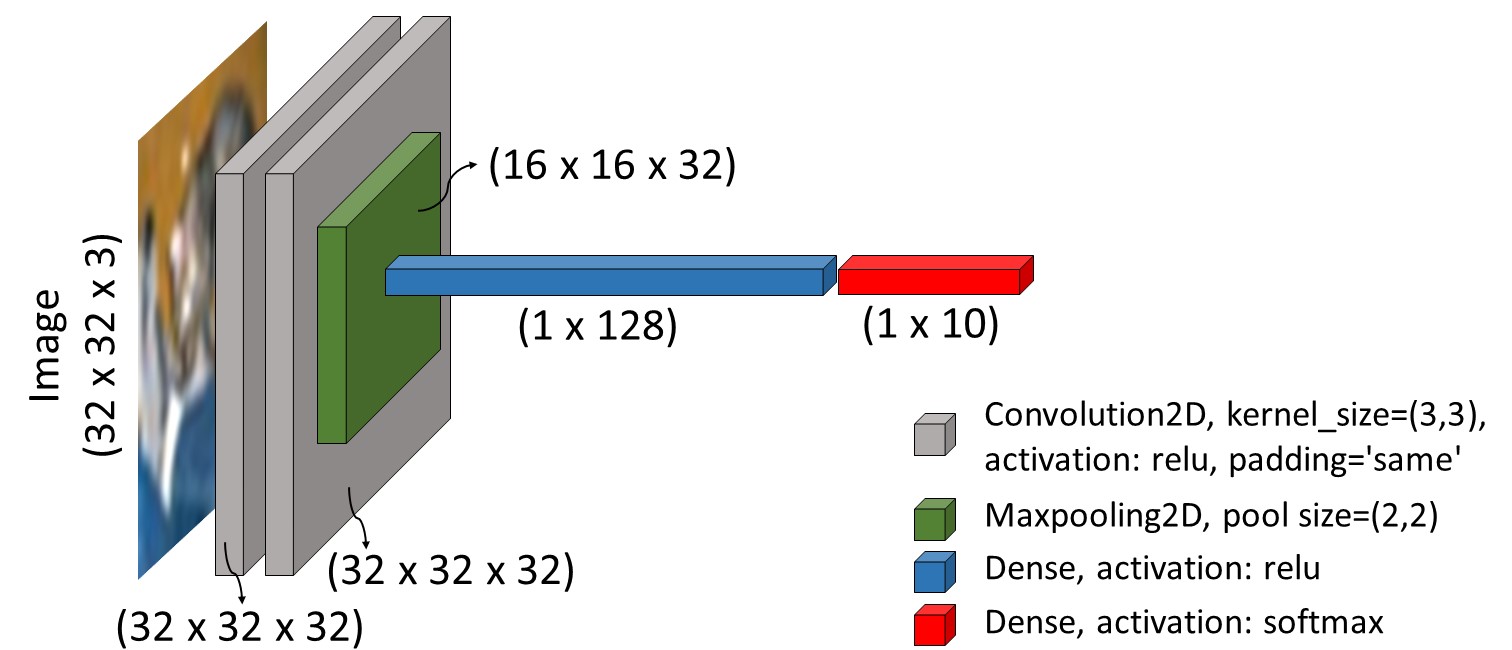}
\caption{Neural network architecture used in our experimental work }
\label{fig:neuralnetwork_fl}
\end{center}
\end{figure*}

\subsection{Experimental Setup}

We implemented a federated learning framework based on Tensorflow\&Keras in Python. In our implementation, there is a server node that keeps and distributes the joint model, and multiple client nodes that update the joint model using their local data. The server and clients have their own data.  The training dataset is distributed among these nodes (including the server) as non-i.i.d. At the beginning, the server sets up a neural network model initialized by training  2 epochs\footnote{epoch means the number of passes over the dataset to carry out learning process} with the data located at the server and then sends this model to the clients. Upon receiving the model from the server, clients train it 2 epochs by using their local data, and then send their locally updated model back to the server. Thus the server collects local models from clients and generates the updated joint model by taking the average of collected model weights. The procedure up to this point is defined as one round in federated learning. After calculating the updated joint model, the server redistributes this updated joint model to clients again and thus initiates another round. By repeating the procedure in this way, the joint model is constantly updated in rounds until being stopped by the server depending on some criteria such as joint model accuracy, convergence level etc. 

We used the CIFAR-10 dataset \cite{CIFAR10krizhevsky2009learning} for experimental analysis. The CIFAR-10 dataset consists of 60000 32x32 color images in 10 different classes, with 6000 images per class, which are split as 50000:10000 for  training and test purposes. The 10 classes in CIFAR-10 represent airplanes, cars, birds, cats, deer, dogs, frogs, horses, ships, and trucks.

\subsection{Malicious Client(s) and Data}

We identify some of the clients as malicious and manipulate a portion of their data to be suitable for performing a backdoor attack. Specifically, we change the labels of the images including a car with a stripe as bird. Thus, similar to \cite{bagdasaryan2020backdoor}, by using such a mislabeled  data in the training process of a malicious client's model, we target to insert a  backdoor into the joint model such that it is supposed to recognize regular car images as car while recognizing the images of cars with stripes as bird.   Figure \ref{fig:backdoordata} shows some samples of such backdoor data.  In our experimental analysis, we  increase the number of malicious clients from  1 (i.e. 10\% of all clients) to 5 (i.e. 50\% of all clients) in different set of experiments.  When there are multiple malicious clients, we distribute malicious data among them equally while keeping  total amount of malicious data the same.%

\subsection{CNN Architecture}

We build a VGG-like \cite{simonyan2014very} CNN architecture for experimental analysis as illustrated in Figure \ref{fig:neuralnetwork_fl}. Because our purpose in this study is not to achieve the best classification performance, rather to investigate abnormal variations in model weights,  we set the CNN architecture simple. In our CNN model, there are two 2D convolution layers with 32 filters, and then a 2D max-pooling layer followed by a fully connected dense layer including 128 neurons, and finally another dense layer with 10 neurons as there are 10 classes in CIFAR10.

\subsection{Results}

In Section \ref{sec:theory} we make a theoretical analysis showing that a backdoor attack carried out  by a malicious user in a federated learning setup may manifest itself with abnormal changes in the final layer's bias values of the local malicious model. To validate this experimentally, we observe deviations in local model weights with respect to the joint model for each training round. In the following set of experiments, the learning rate is set to 0.01 unless otherwise stated.

\subsubsection{Varying Number of Malicious Clients}
\label{sec:VaryingNumberMaliciousClients}

\begin{figure*}[h!]
\captionsetup[subfigure]{justification=centering}
    \centering
    \begin{subfigure}[t]{0.32\textwidth}
        \includegraphics[width=\textwidth]{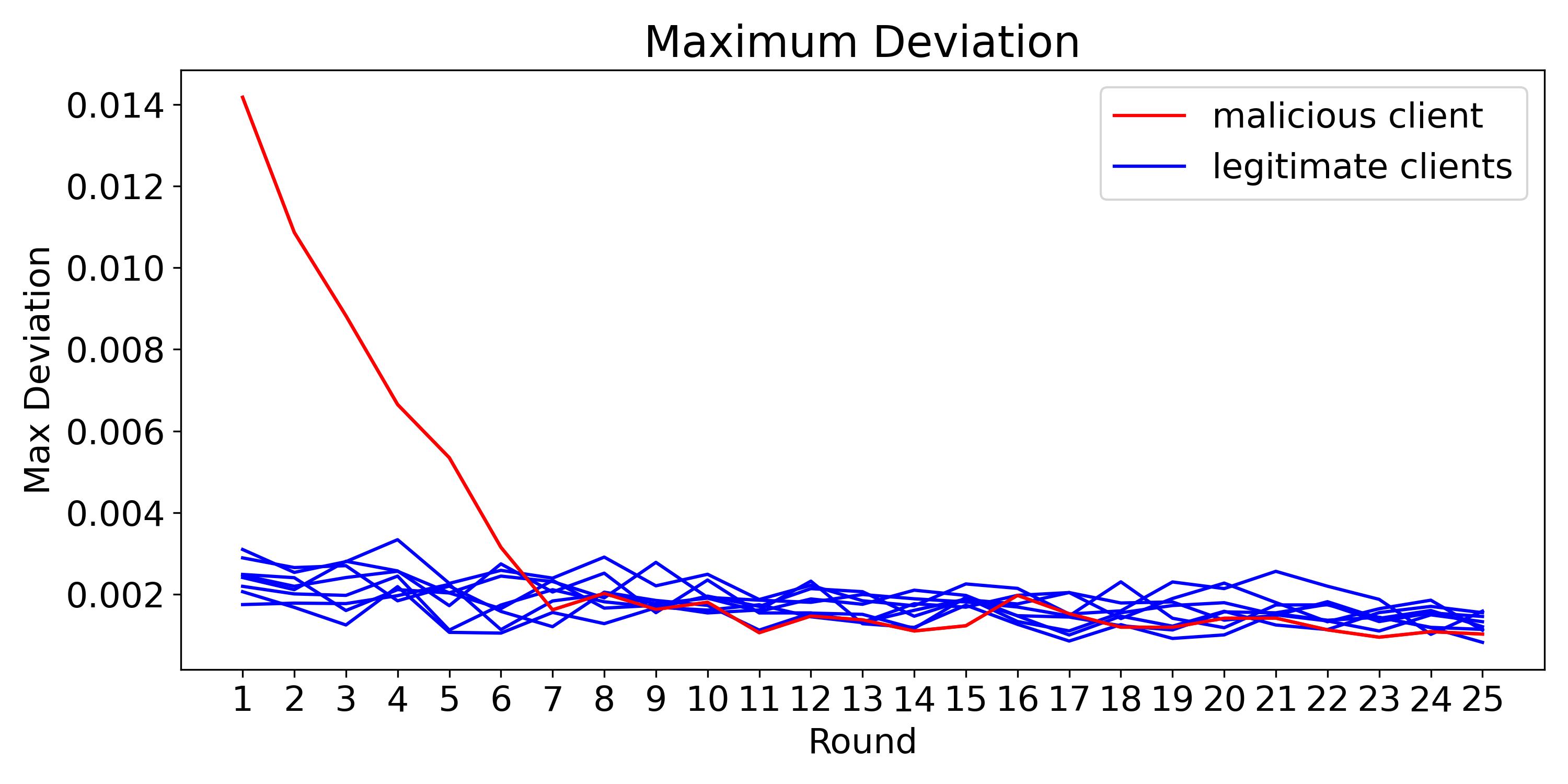}
    \end{subfigure}
    \begin{subfigure}[t]{0.32\textwidth}
        \includegraphics[width=\textwidth]{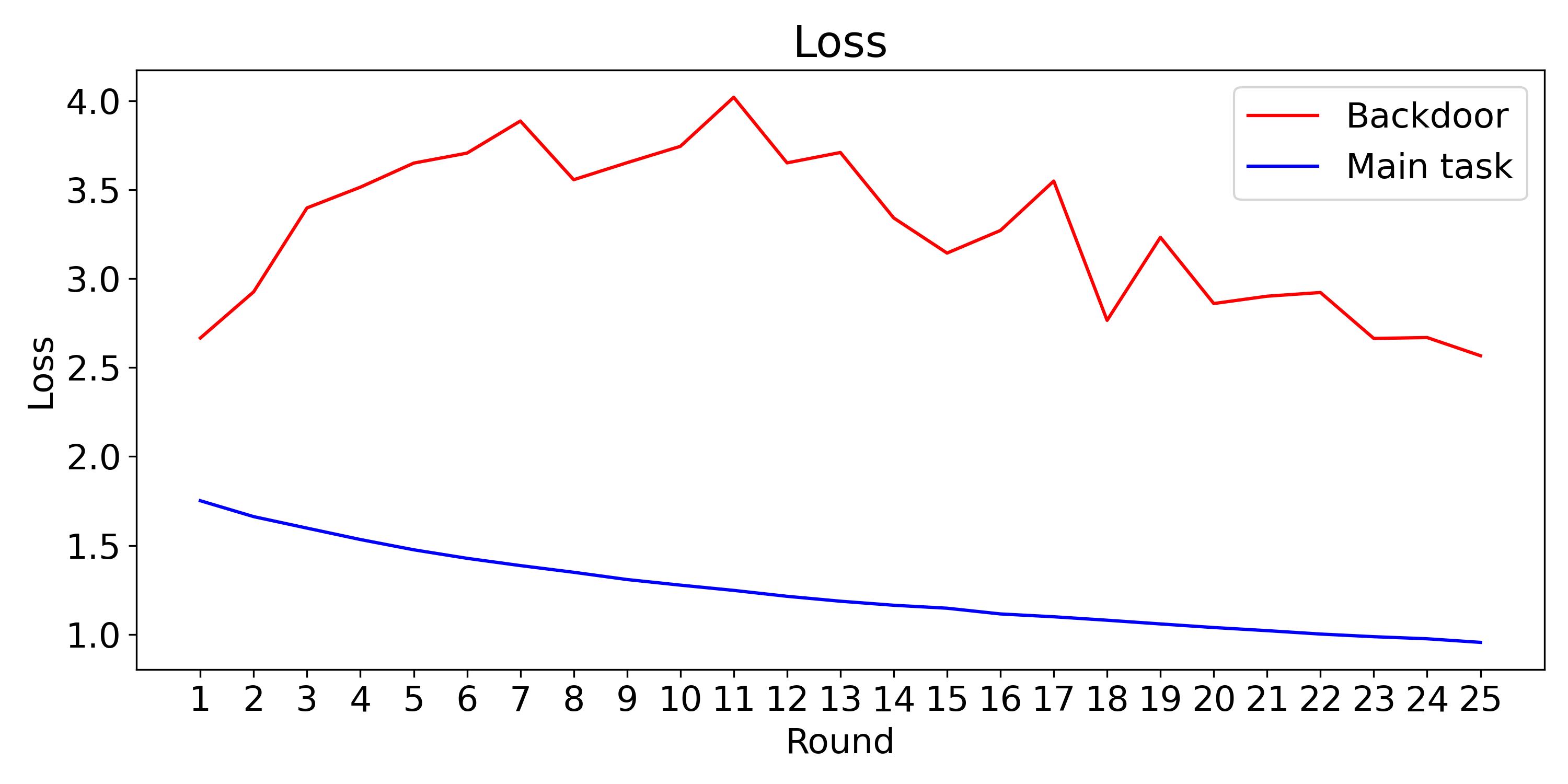}
        \caption{1 malicious client}
        \label{fig:1malicious_a}
    \end{subfigure}
    \begin{subfigure}[t]{0.32\textwidth}
        \includegraphics[width=\textwidth]{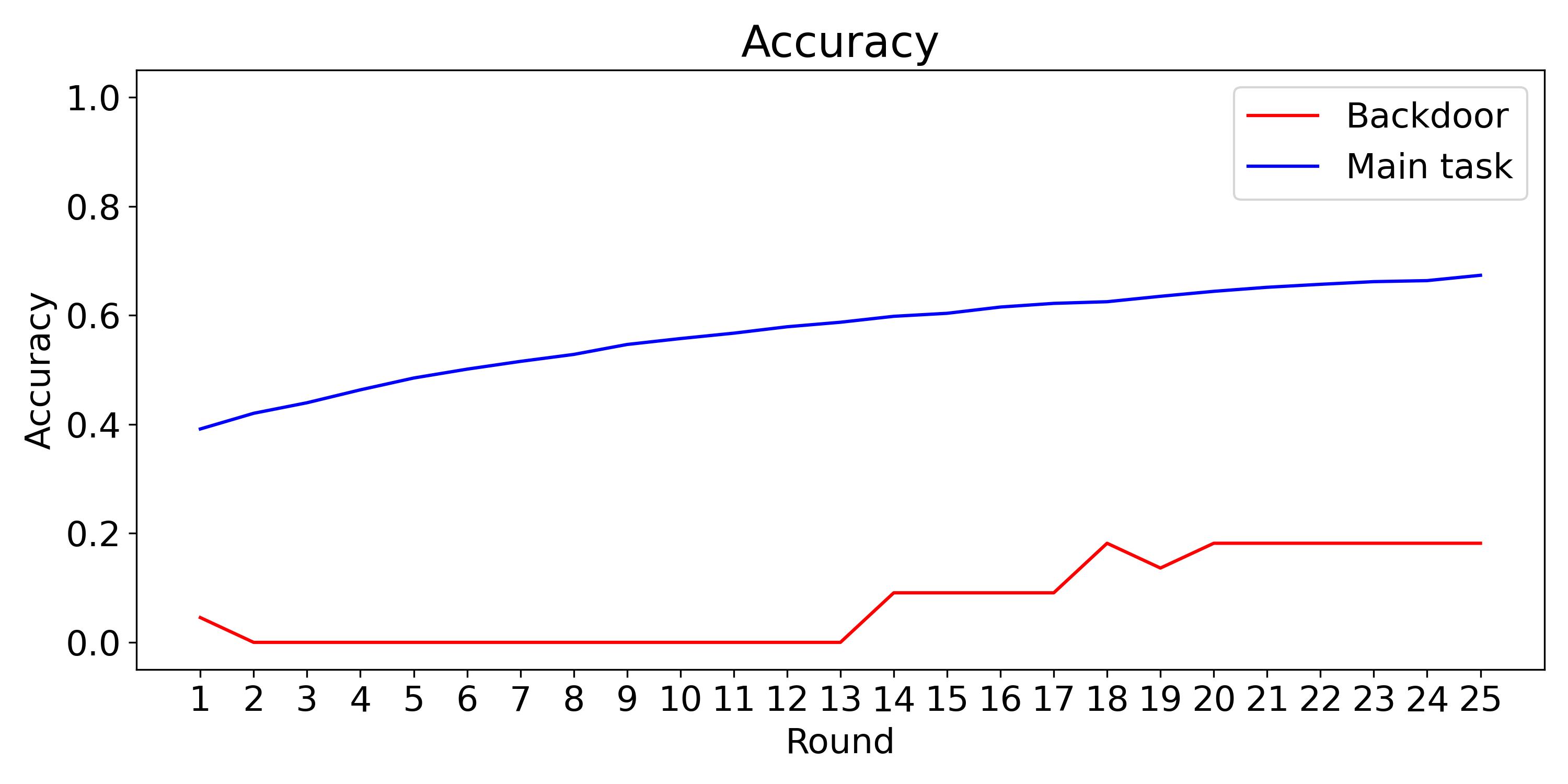}
    \end{subfigure}
    \begin{subfigure}[t]{0.32\textwidth}
        \includegraphics[width=\textwidth]{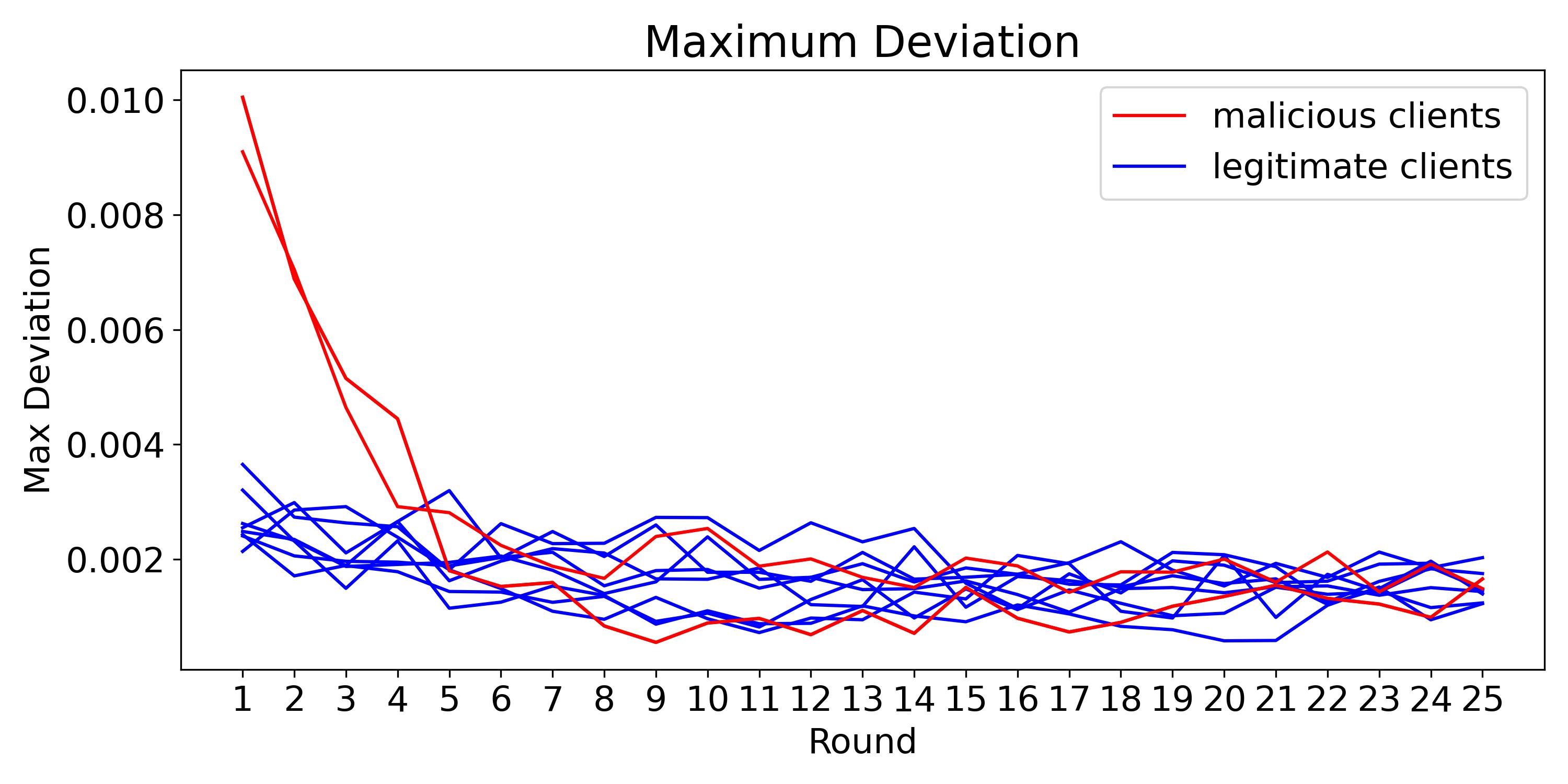}
    \end{subfigure}
    \begin{subfigure}[t]{0.32\textwidth}
        \includegraphics[width=\textwidth]{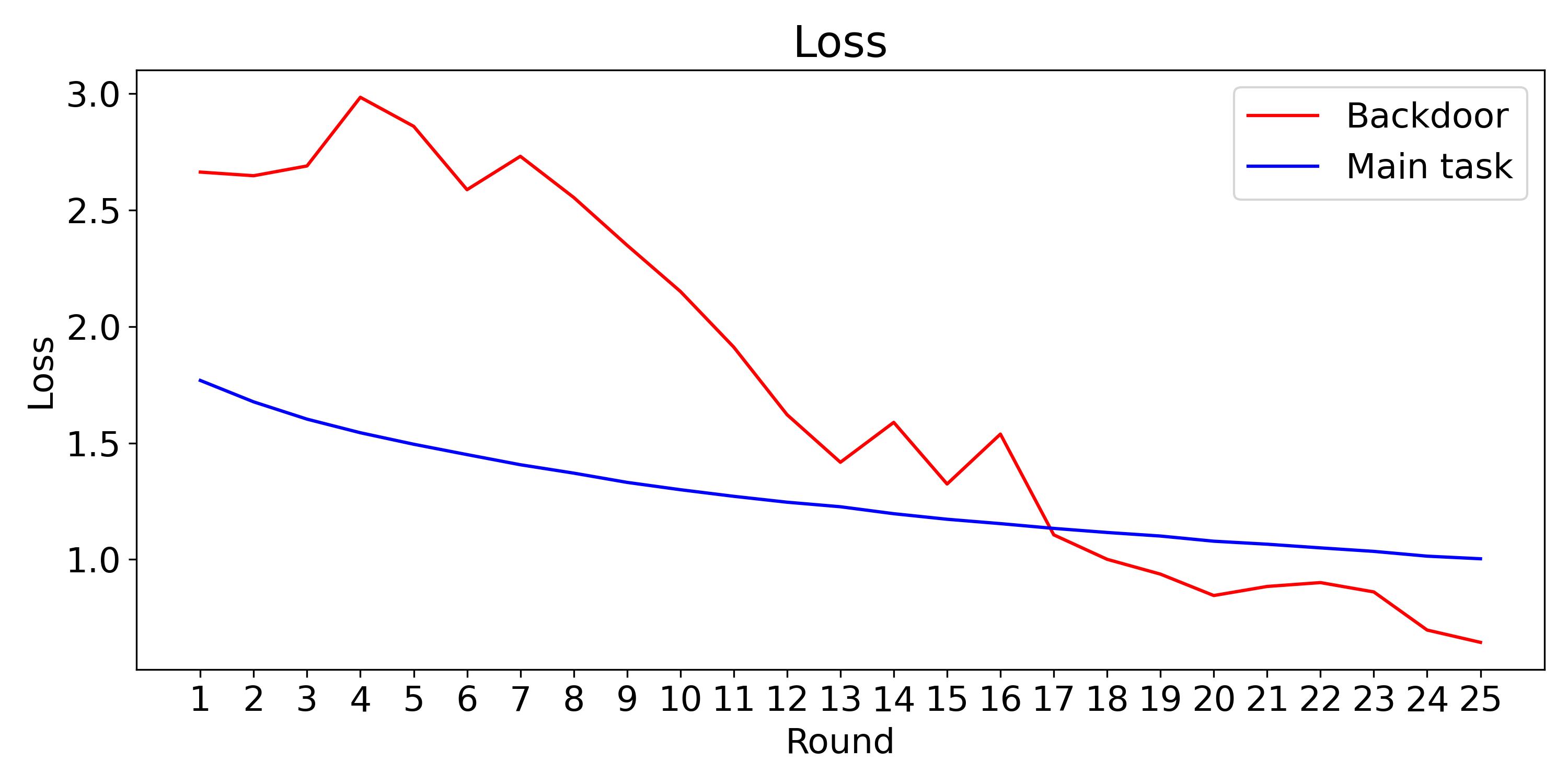}
        \caption{2 malicious clients}
        \label{fig:1malicious_b}
    \end{subfigure}
    \begin{subfigure}[t]{0.32\textwidth}
        \includegraphics[width=\textwidth]{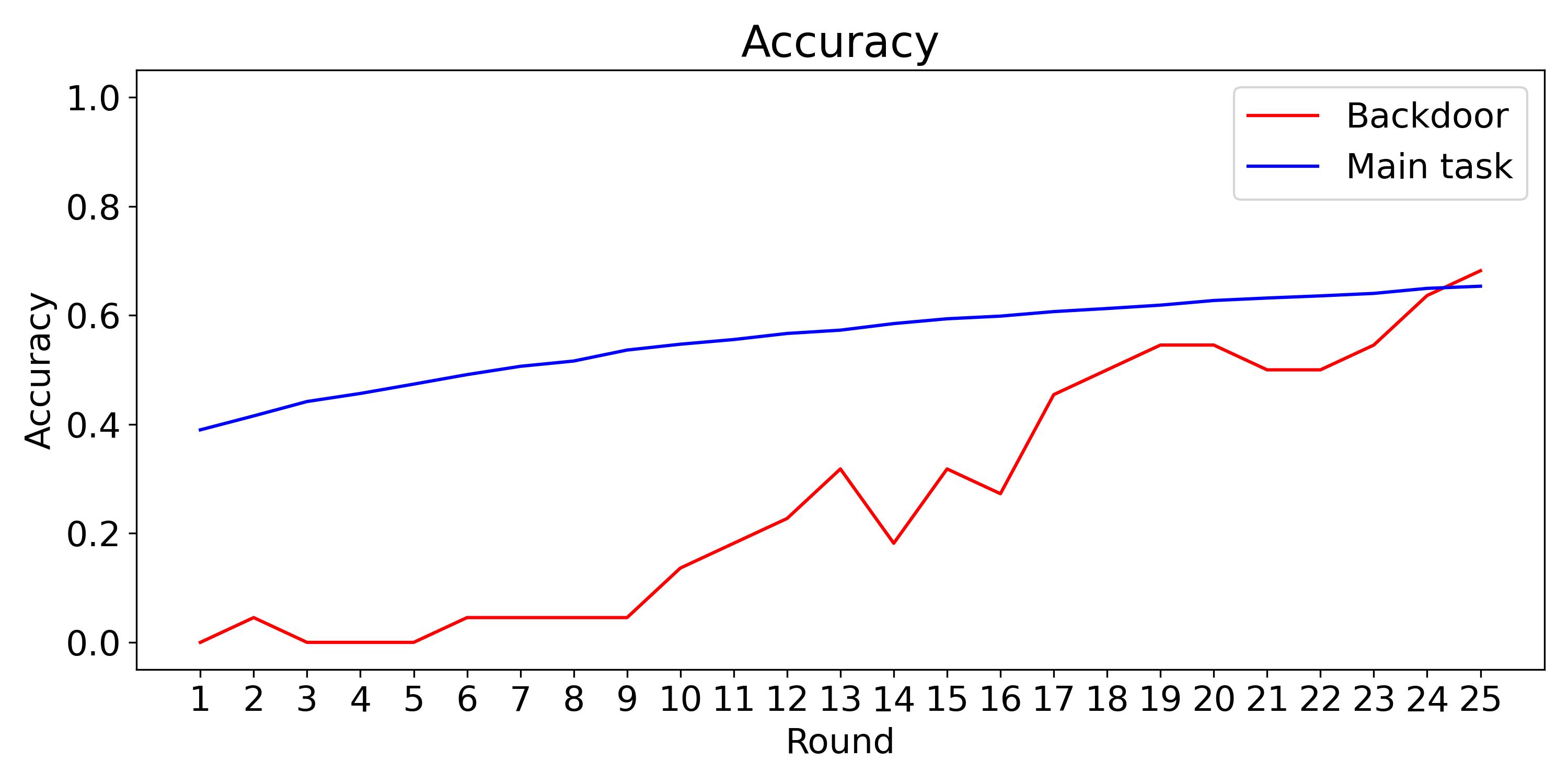}
    \end{subfigure}
    \begin{subfigure}[t]{0.32\textwidth}
        \includegraphics[width=\textwidth]{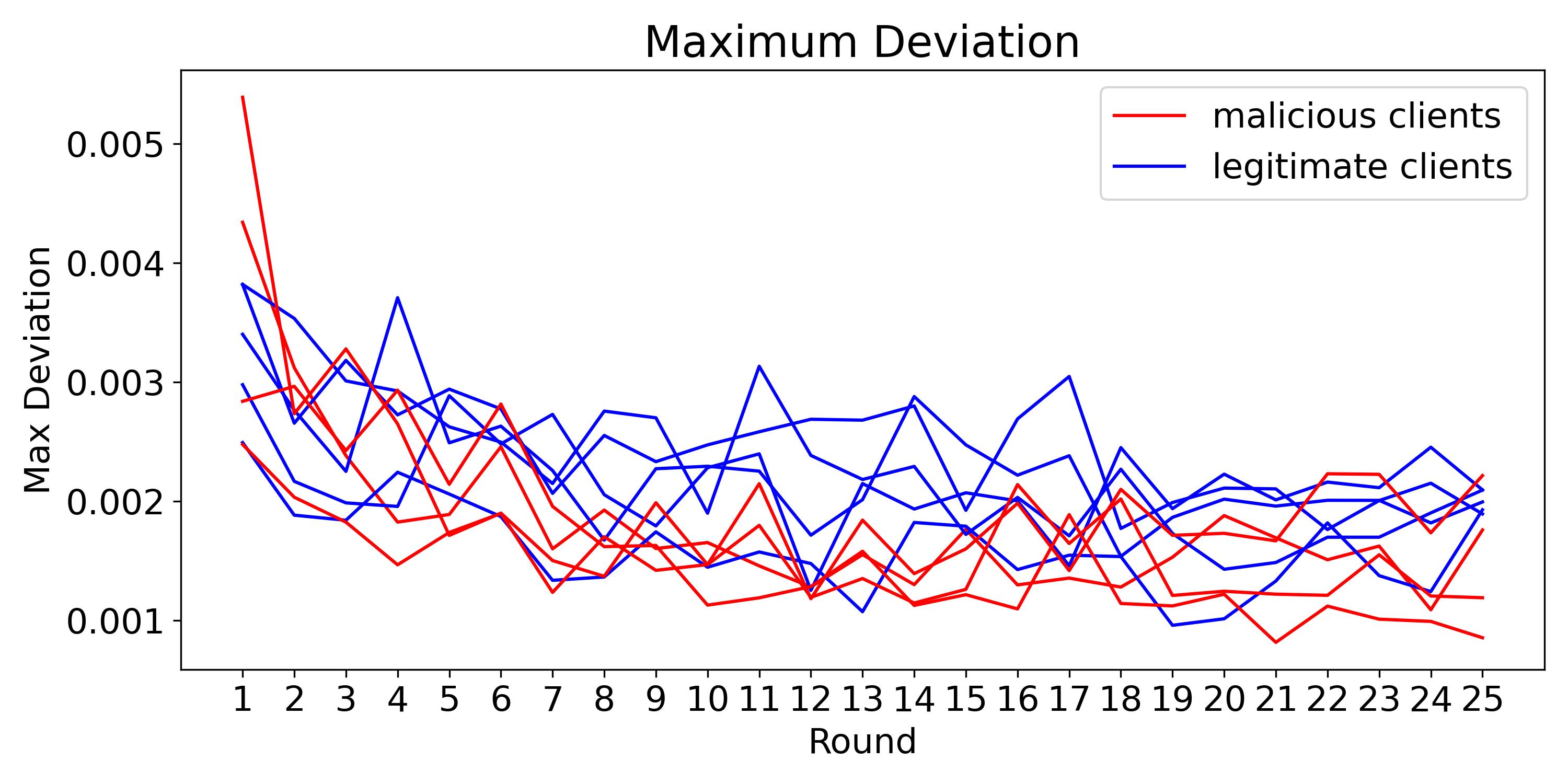}
    \end{subfigure}
    \begin{subfigure}[t]{0.32\textwidth}
        \includegraphics[width=\textwidth]{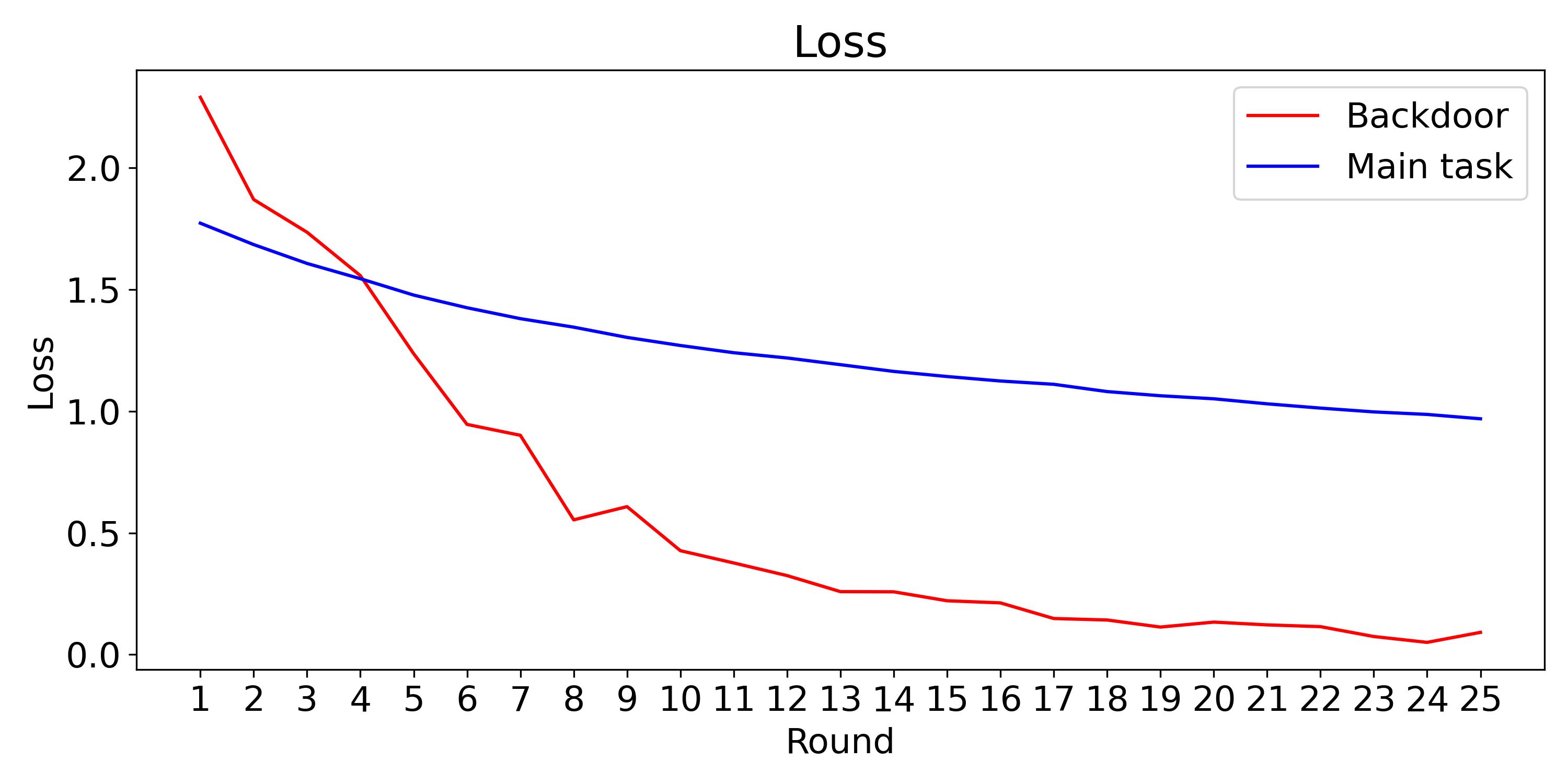}
        \caption{5 malicious clients}
        \label{fig:1malicious_c}
    \end{subfigure}
    \begin{subfigure}[t]{0.32\textwidth}
        \includegraphics[width=\textwidth]{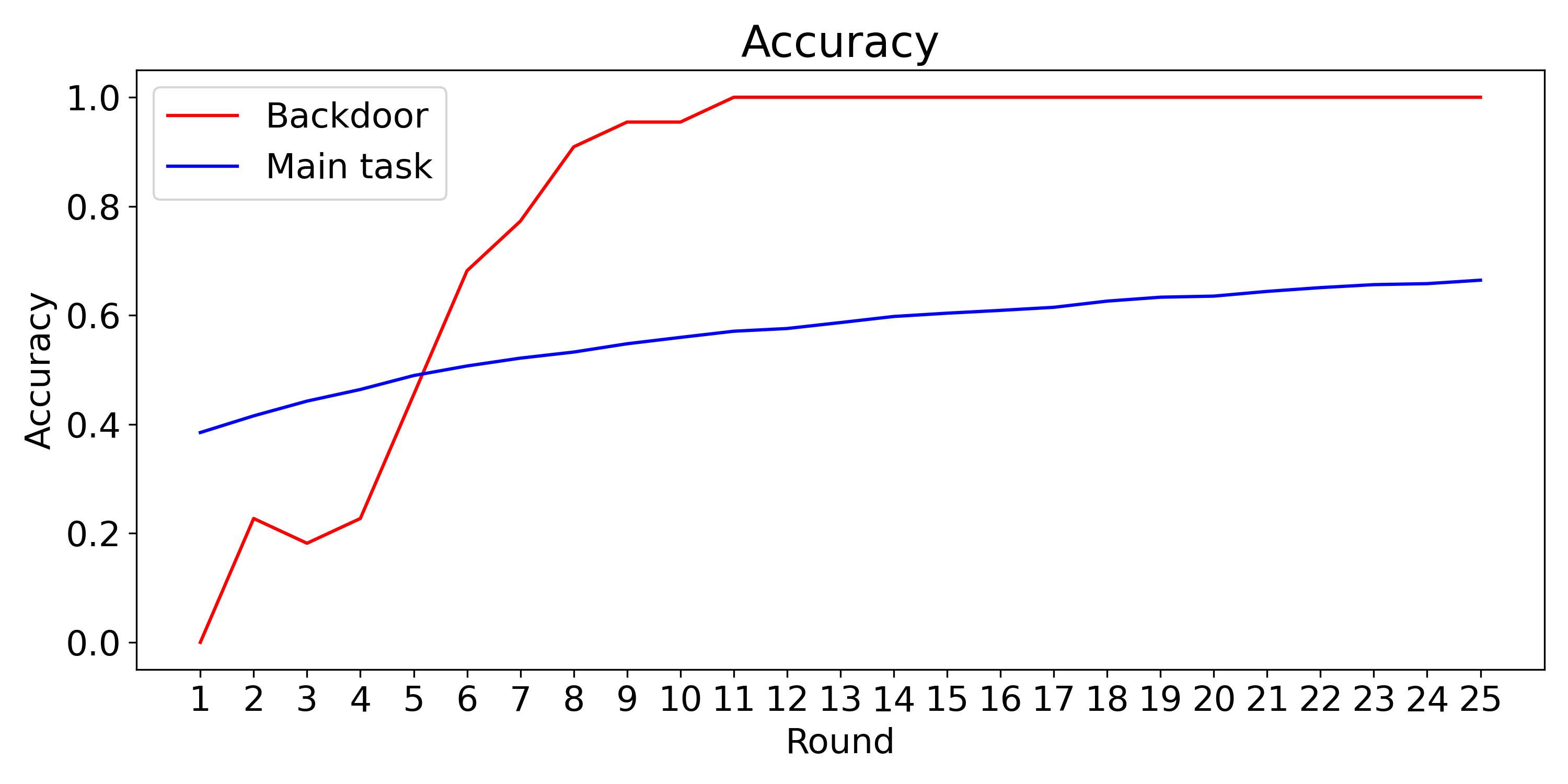}
    \end{subfigure}    
    \caption{Results for the cases including  (a) 1 malicious client, (b) 2 malicious clients, and (c) 5 malicious clients out of 10 clients.}
    \label{fig:1maliciouscase1}
\end{figure*}

In our first experimental setup, there are 10 clients, one of which is malicious (i.e. the client no 1),  and we observe the maximum deviation in the updated local model weights with respect to the distributed joint model weights in all training rounds. In the malicious client, we adjust the backdoored data rate to 1/3 by generating copies of the malicious samples. In our investigation of the experimental results, we have not observed any abnormal deviations in the collected local model weights except the final layer bias values of the malicious client's model. Figure \ref{fig:1malicious_a} shows our experimental results for the maximum deviation in the final layer bias values of local models as well as joint model loss  and joint model accuracy for both main task and backdoor  up to the 25th training round. As seen in Figure \ref{fig:1malicious_a}, the maximum deviation curve of the malicious client differs clearly from other curves at the beginning of training by showing relatively larger values, and the magnitude of this abnormality gradually decreases until the 7th training round, after which the malicious model acts normal like the others. Such an abnormal behavior of the malicious client is indeed consistent with our theoretical analysis in Section \ref{sec:theory} and particularly validates our justification regarding Equation \ref{eq:finalall} there. But why does the maximum deviation curve of the malicious client gradually decrease as training proceeds? Firstly, this is partly because of the fact that the malicious local model and the joint model  get more converged each other as training proceeds in rounds. Although this convergence is not enough to effectively insert the backdoor into the joint model as seen in the Loss and Accuracy plots of Figure \ref{fig:1malicious_a} where backdoor loss is relatively high and backdoor accuracy seems very low respectively, the maximum deviation due to malicious data still decreases as the joint model is updated by collected local models. Furthermore, since the training process takes place in batches of data, gradients in backpropagation  are calculated as average over all samples in a batch that contains both malicious and legitimate data in a malicious client given that the malicious data rate is 1/3. Therefore, the maximum deviation decreases as joint model accuracy on the main task increases even if joint model accuracy on the backdoor does not increase.

In our second experimental setup, we include 2 malicious clients out of 10 clients and distribute copies of malicious samples equally  among them so that the backdoored sample rate on each malicious client becomes 1/6 as keeping the amount of cumulative malicious data same. In this experiment, similar to the previous experiment, we measure the maximum deviations in local model updates as well as joint model loss and accuracy in consecutive training rounds. Figure \ref{fig:1malicious_b} shows our experimental results in this experiment. According to Figure \ref{fig:1malicious_b}, the malicious models indicate relatively larger maximum deviation values with respect to the legitimate clients in the first training rounds, and then  their deviations gradually  decrease in the following rounds and reach the levels that cannot be differentiated from the legitimate models. This is similar to the malicious client behavior observed in the previous experiment, but with the difference that (i) the malicious local models and the joint model get converged faster here at around training round 5 (w.r.t. round 7 in the previous one-malicious client case), and (ii) two malicious clients seem to be more successful at inserting a backdoor into the joint model as seen in the Loss and Accuracy plots of Figure \ref{fig:1malicious_b}. Put it differently, this experiment indicates that two malicious clients carry out more effective attacks with respect to one-malicious client case even though the amount of cumulative malicious data is identical in either case.

In our third experimental setup, we increase the number of malicious clients to 5, which is equivalent to 50\% of all clients, while keeping the amount of cumulative malicious data the same,  and repeat the earlier experiments.  Figure \ref{fig:1malicious_c}  shows the maximum deviations in the final layer's bias values as well as Loss and Accuracy curves  in 25 consecutive training rounds for this case. As seen in Figure \ref{fig:1malicious_c}, the abnormal behavior of  the malicious clients is no longer apparent in the maximum deviation curves as all the blue and red lines are intertwined in all rounds. This is intuitive because the malicious clients are no longer in the minority and therefore their behavior is not perceived as an anomaly. As seen in the Loss and Accuracy plots of  Figures \ref{fig:1malicious_c}, joint model loss and accuracy for the backdoor reach better values with respect to the main task, which indicates that the number of malicious clients is an important factor in the effectiveness of the attack.  

In addition to the above interpretations, we can also infer from the experimental results given in this part that  malicious model updates are likely to display abnormal deviations during the attempt of inserting a backdoor, whereas they no longer indicate such anomaly during the attempts of maintaining an already inserted backdoor, which implies that it becomes critical to detect such anomalies in the early stages of backdoor insertion.



\subsubsection{Impact of Learning Rate }

\begin{figure*}[h!]
    \centering
    \begin{subfigure}[t]{0.32\textwidth}
        \includegraphics[width=\textwidth]{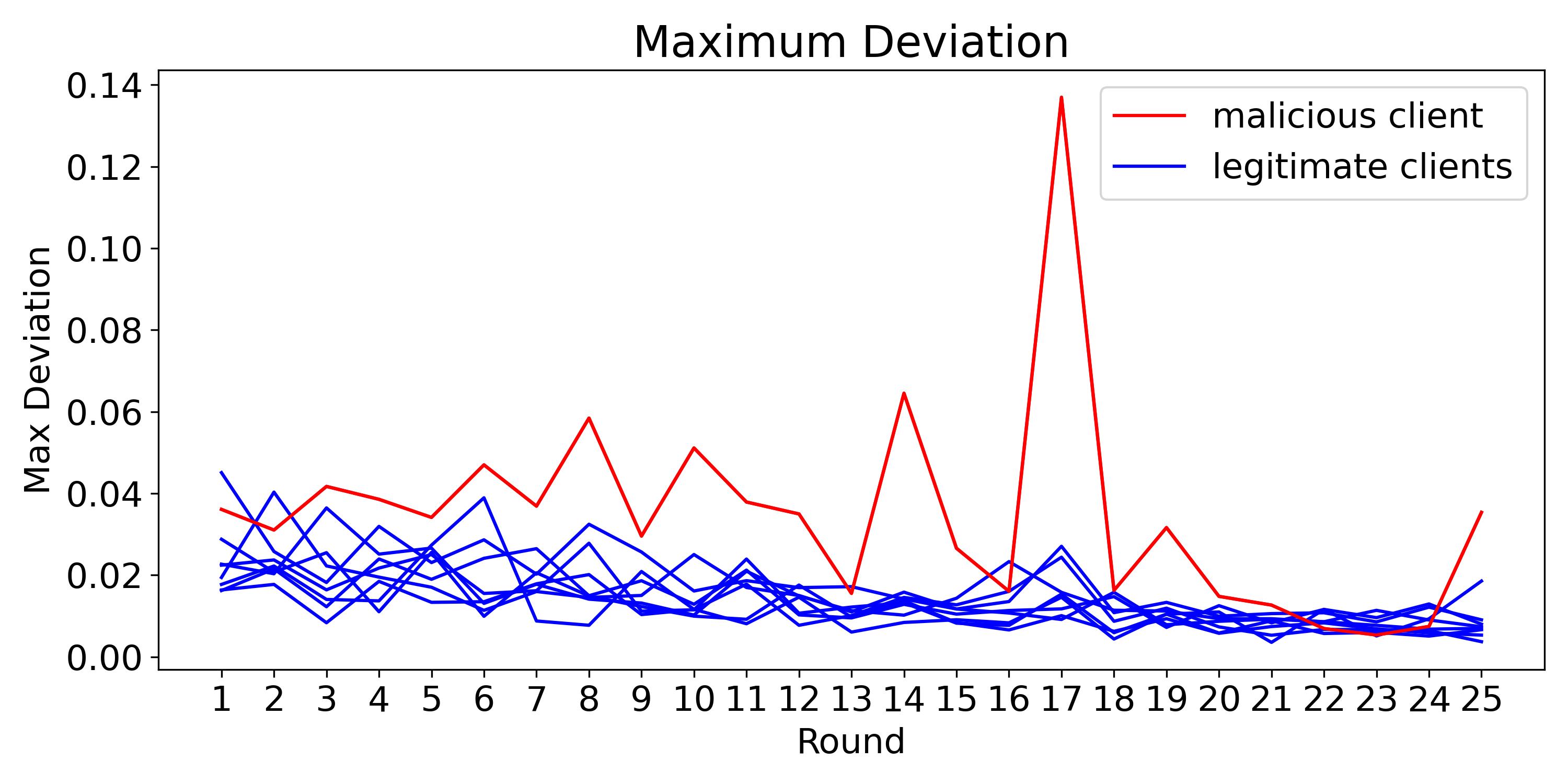}
    \end{subfigure}
    \begin{subfigure}[t]{0.32\textwidth}
        \includegraphics[width=\textwidth]{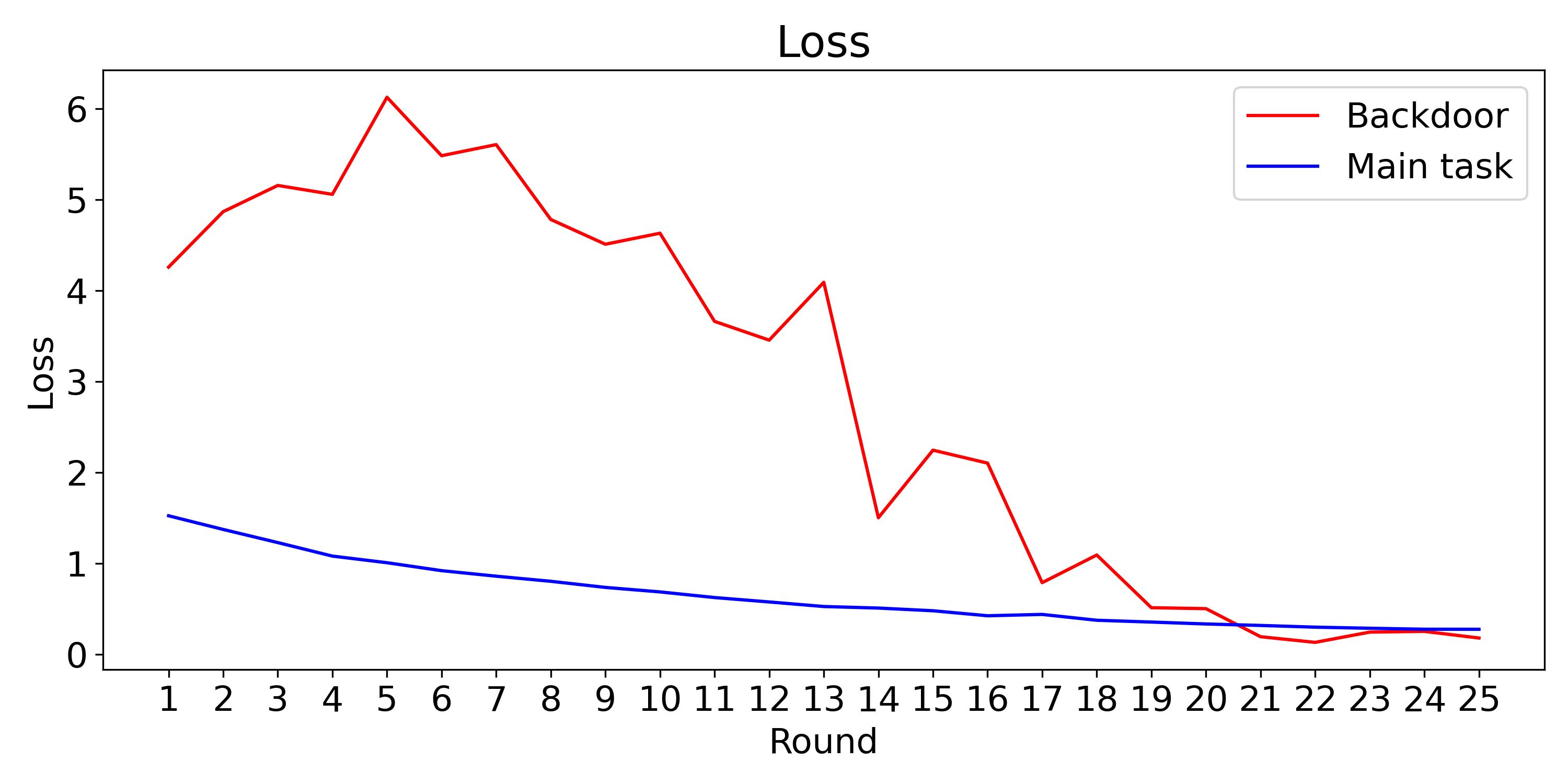}
        \caption{Learning rate = 0.1}
        \label{fig:lr_a}
    \end{subfigure}
    \begin{subfigure}[t]{0.32\textwidth}
        \includegraphics[width=\textwidth]{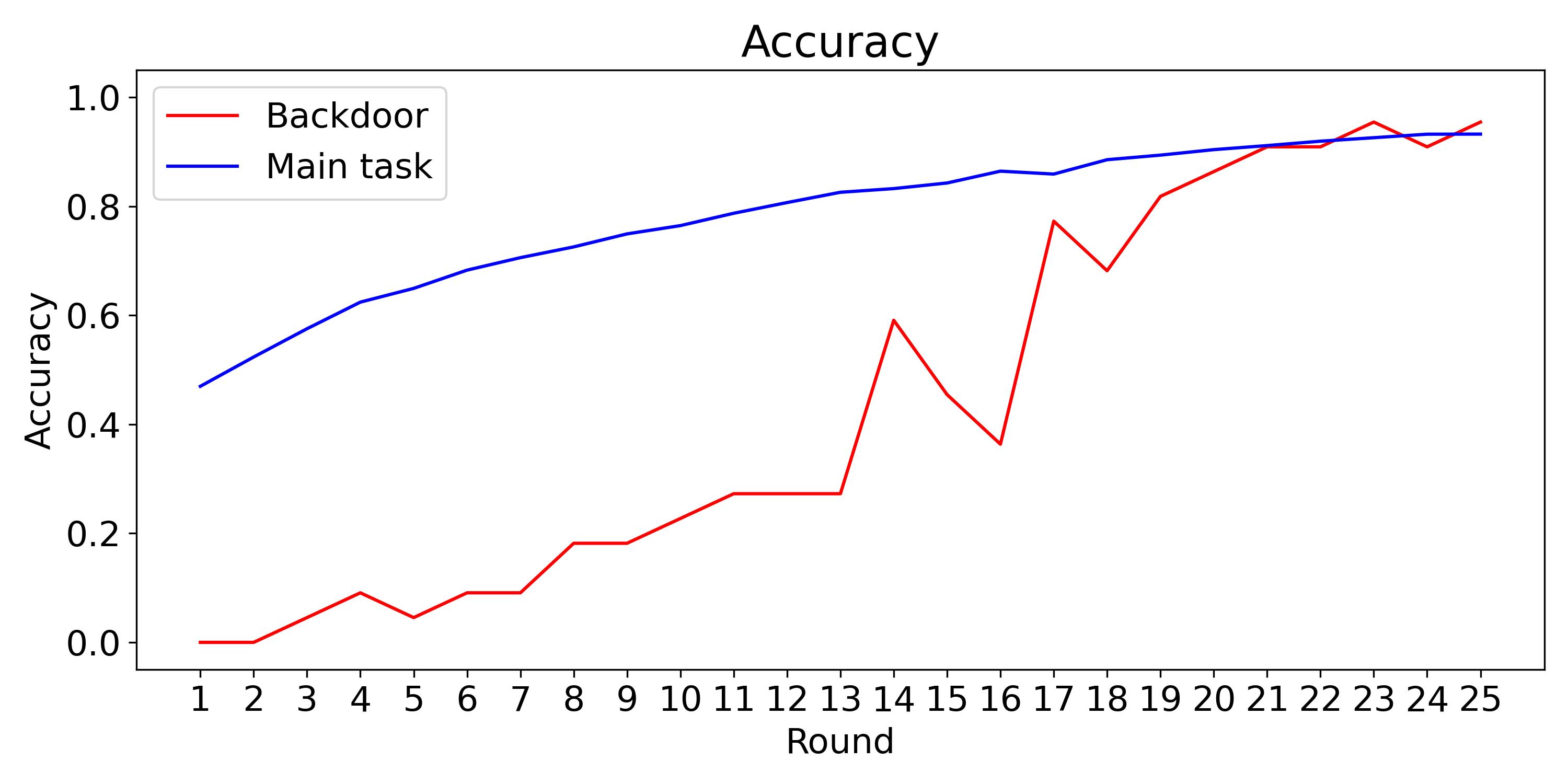}
    \end{subfigure}
    \begin{subfigure}[t]{0.32\textwidth}
        \includegraphics[width=\textwidth]{figures/maliciousclientnumber/maxdeviation20211112_1malicious.jpg}
    \end{subfigure}
    \begin{subfigure}[t]{0.32\textwidth}
        \includegraphics[width=\textwidth]{figures/maliciousclientnumber/Loss20211112_1malicious.jpg}
        \caption{Learning rate = 0.01}
        \label{fig:lr_b}
    \end{subfigure}
    \begin{subfigure}[t]{0.32\textwidth}
        \includegraphics[width=\textwidth]{figures/maliciousclientnumber/Accuracy20211112_1malicious.jpg}
    \end{subfigure}
    \begin{subfigure}[t]{0.32\textwidth}
        \includegraphics[width=\textwidth]{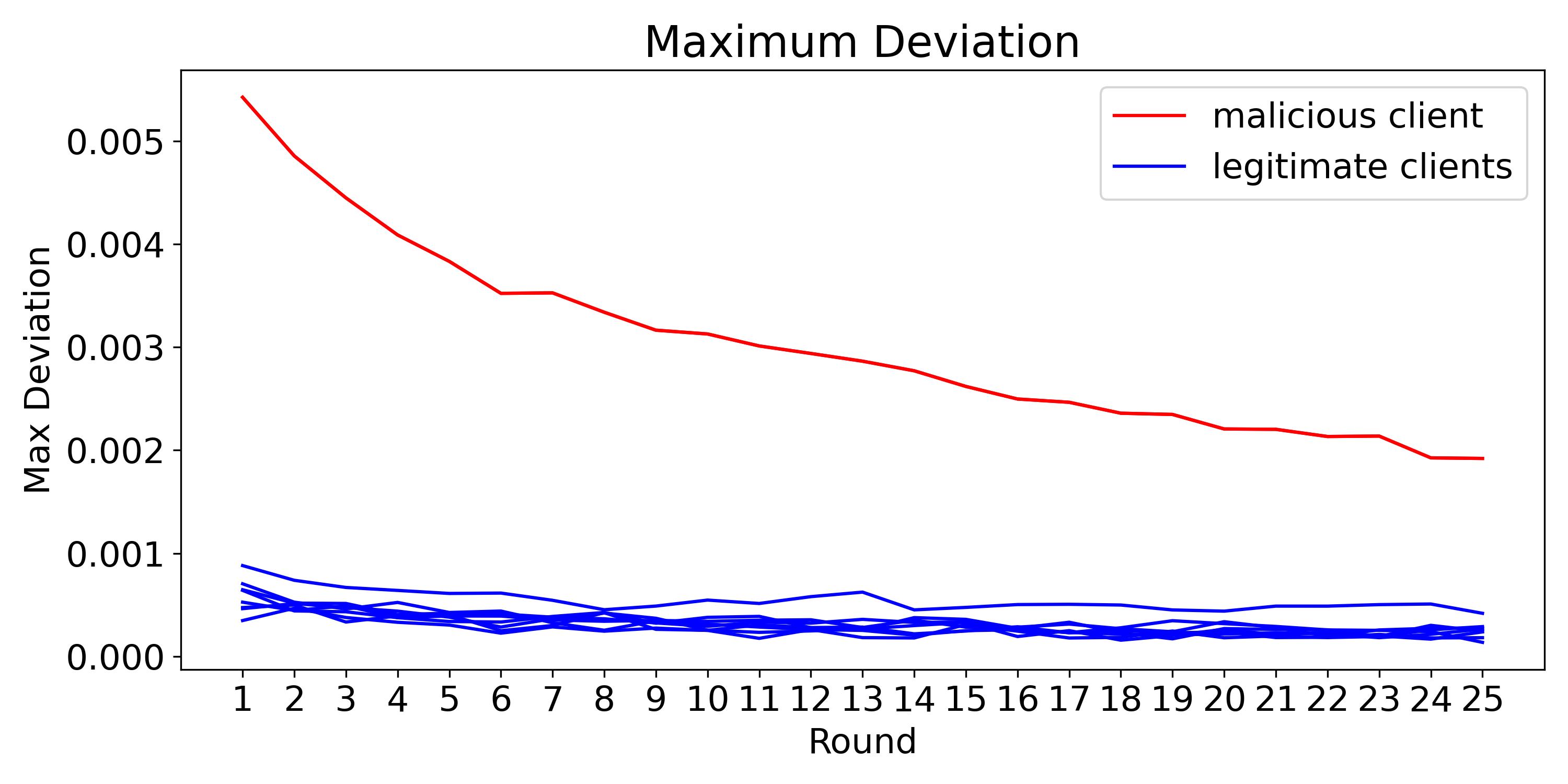}
    \end{subfigure}
    \begin{subfigure}[t]{0.32\textwidth}
        \includegraphics[width=\textwidth]{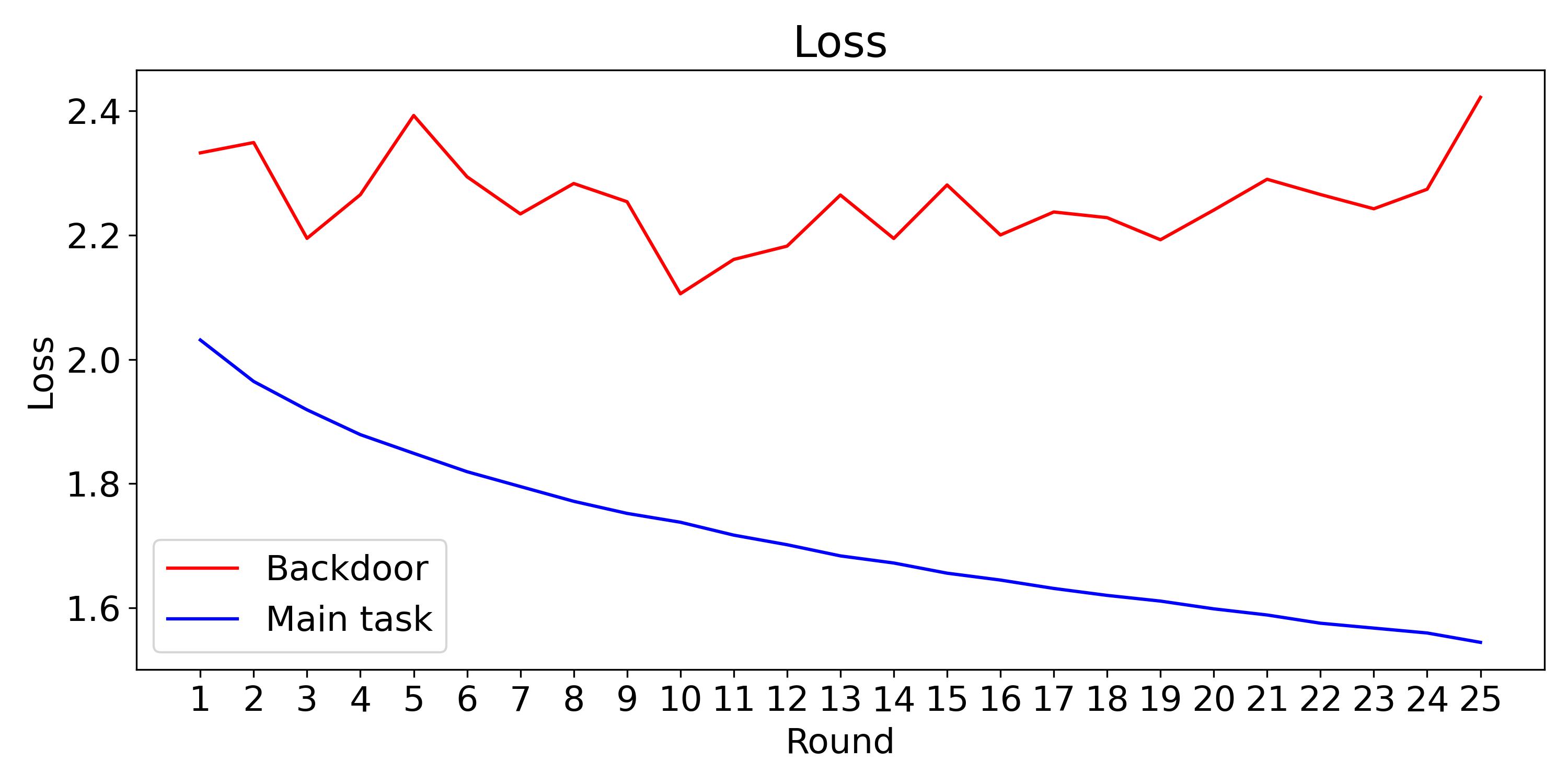}
        \caption{Learning rate = 0.001}
        \label{fig:lr_c}
    \end{subfigure}
    \begin{subfigure}[t]{0.32\textwidth}
        \includegraphics[width=\textwidth]{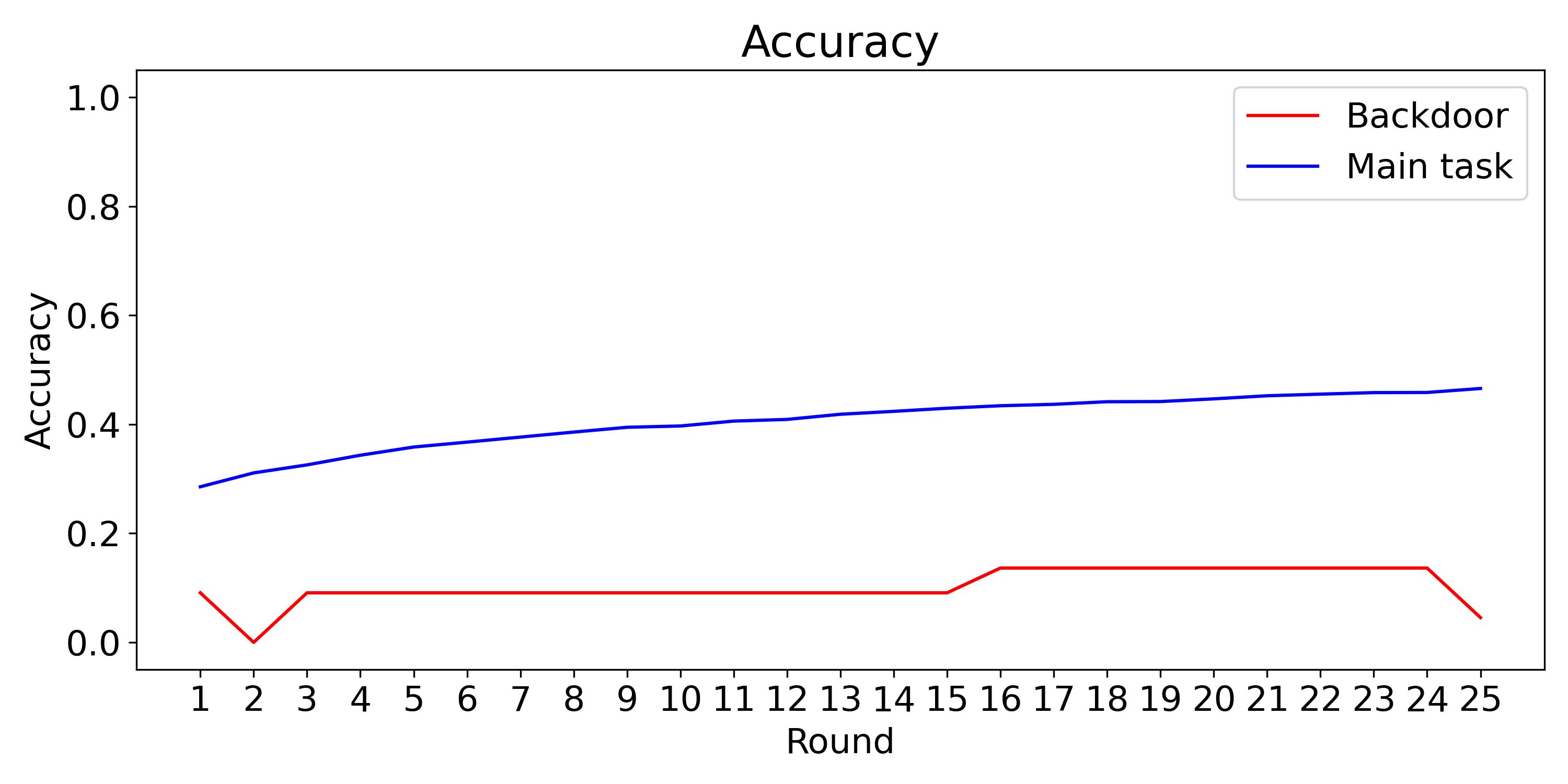}
    \end{subfigure}
    \caption{Impact of different learning rates of (a) 0.1, (b) 0.01, and (c) 0.001 on the maximum deviation in the last layer bias values, joint model loss and accuracy for both main task and backdoor in consecutive training rounds.}
    \label{fig:learningrate}
\end{figure*}

In this set of experiments, we deepen our investigation using different learning rates while involving one malicious client with the malicious data rate of 1/3. Specifically, we observe the maximum deviation in the final layer's bias values of the local models as well as Loss and Accuracy of the joint model for both the main task and the backdoor in consecutive training rounds.  Figure \ref{fig:learningrate} depicts our experimental results for three different learning rates of 0.1, 0.01, and 0.001. 

Figure \ref{fig:lr_a} includes our experimental results for the learning rate of 0.1. As seen in  Figure \ref{fig:lr_a}, the malicious client indicates abnormal deviations with sharp ups and downs in most training rounds, which happens most likely due to (i) high learning rate because  the amount of change in the weights during model training is directly proportional to learning rate according to Equations \ref{eq:overall1}{\&}\ref{eq:overall2}, and (ii) federated learning where the malicious client and legitimate clients compete with each other to reflect their influence on the joint model. More specifically, when the backdoor loss reaches a minimal level in the joint model, the malicious client makes relatively less amount of contribution for backdoor in the following rounds, which paves the way for legitimate clients to dominate causing sudden changes in the direction of increasing backdoor loss. When backdoor loss increases, malicious client's impact on the joint model increases again in proportional with learning rate and loss,   
and the cycle continues like this until convergence occurs. This pattern is seen in loss and accuracy plots of Figure \ref{fig:lr_a}. We also see from the accuracy plot of Figure \ref{fig:lr_a} that the main task and backdoor accuracy can simultaneously reach maximum accuracy level (i.e. 100\%), which means that the backdoor is successfully inserted without degrading the model's performance on the main task. 

Figure \ref{fig:lr_b} includes our experimental results for the learning rate of 0.01. This is the same experimental results given in  Figure \ref{fig:1malicious_a}, and therefore the interpretations made earlier for Figure \ref{fig:1malicious_a} is also valid for Figure \ref{fig:lr_b}.     

Figure \ref{fig:lr_c} includes our experimental results for the learning rate of 0.001. The malicious client continues to demonstrate abnormal deviations in this case as seen in  Figure \ref{fig:lr_c}, yet the scale of the deviation is different with respect to other learning rates, which is intuitive due to the limiting role of learning rate on the strength of updates as discussed above. 

When we comparatively examine all the results given in Figure \ref{fig:learningrate} regarding three different learning rates, we can see that the change in the joint model occurs at a rate proportional to the learning rate, which is intuitive.


\subsubsection{Impact of Malicious Data Rate}

\begin{figure*}[h!]
    \centering
    \begin{subfigure}[t]{0.32\textwidth}
        \includegraphics[width=\textwidth]{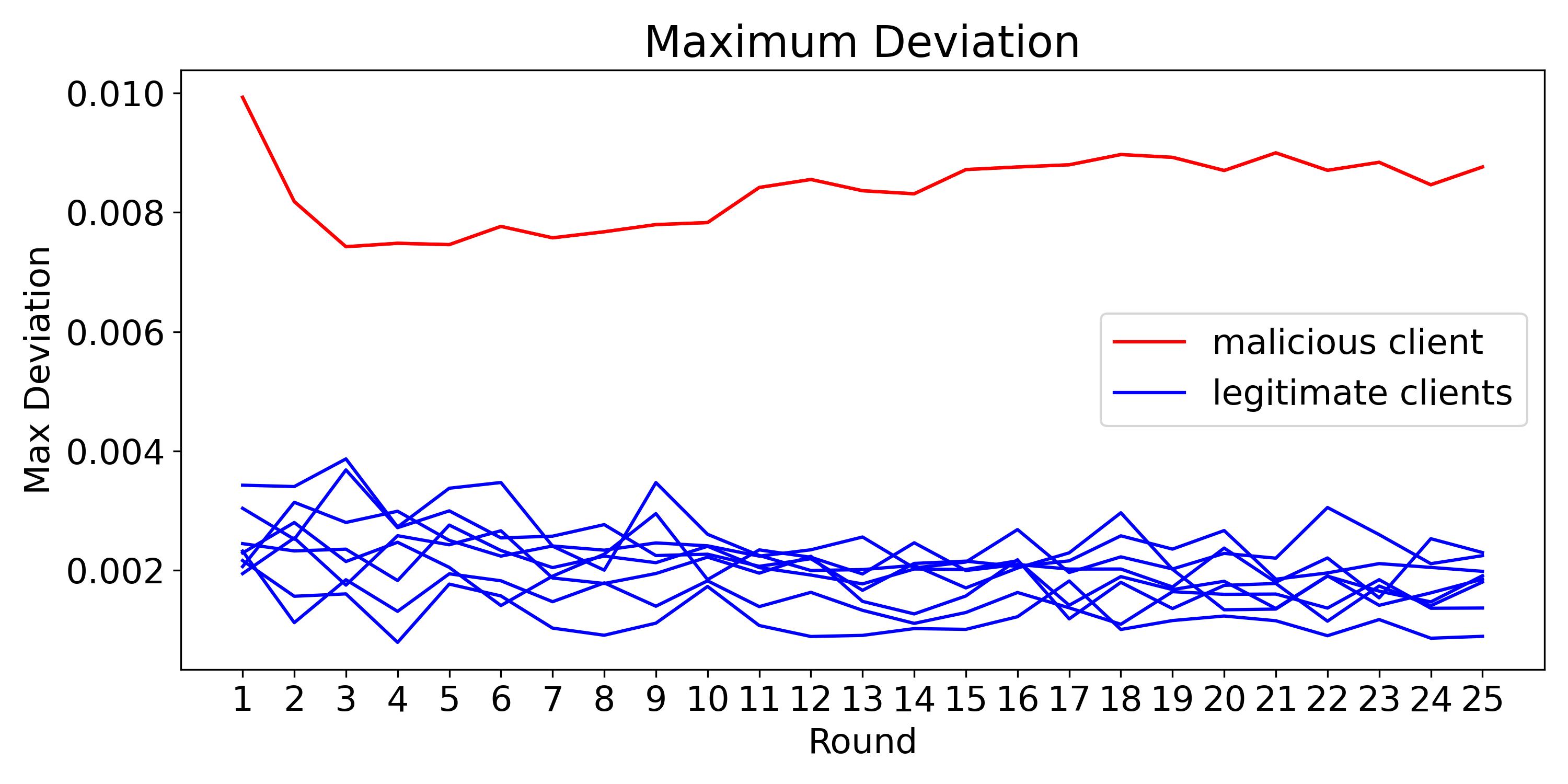}
    \end{subfigure}
    \begin{subfigure}[t]{0.32\textwidth}
        \includegraphics[width=\textwidth]{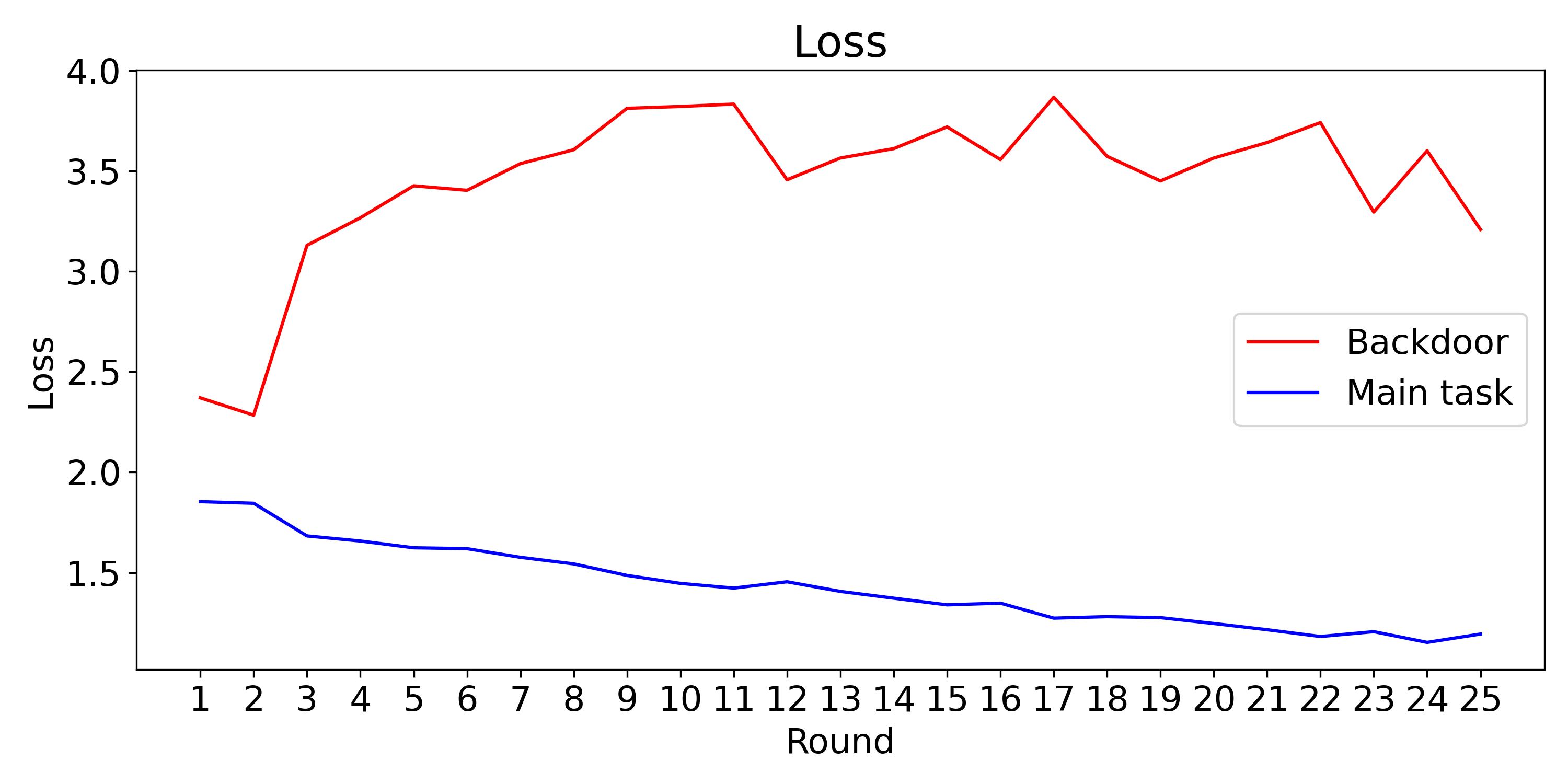}
        \caption{Malicious data rate = 1/1}
        \label{fig:mdr_a}
    \end{subfigure}
    \begin{subfigure}[t]{0.32\textwidth}
        \includegraphics[width=\textwidth]{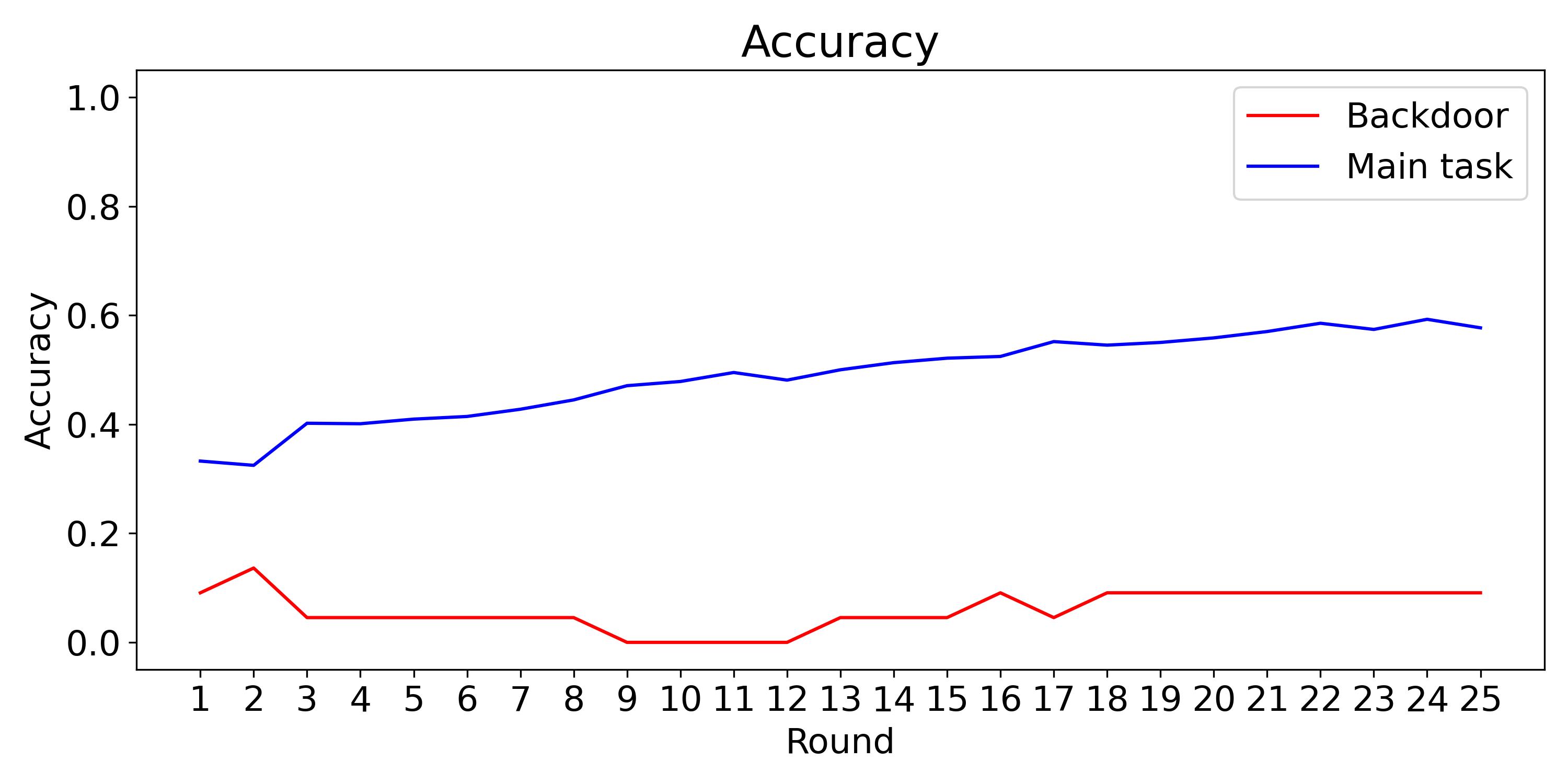}
    \end{subfigure}
    \begin{subfigure}[t]{0.32\textwidth}
        \includegraphics[width=\textwidth]{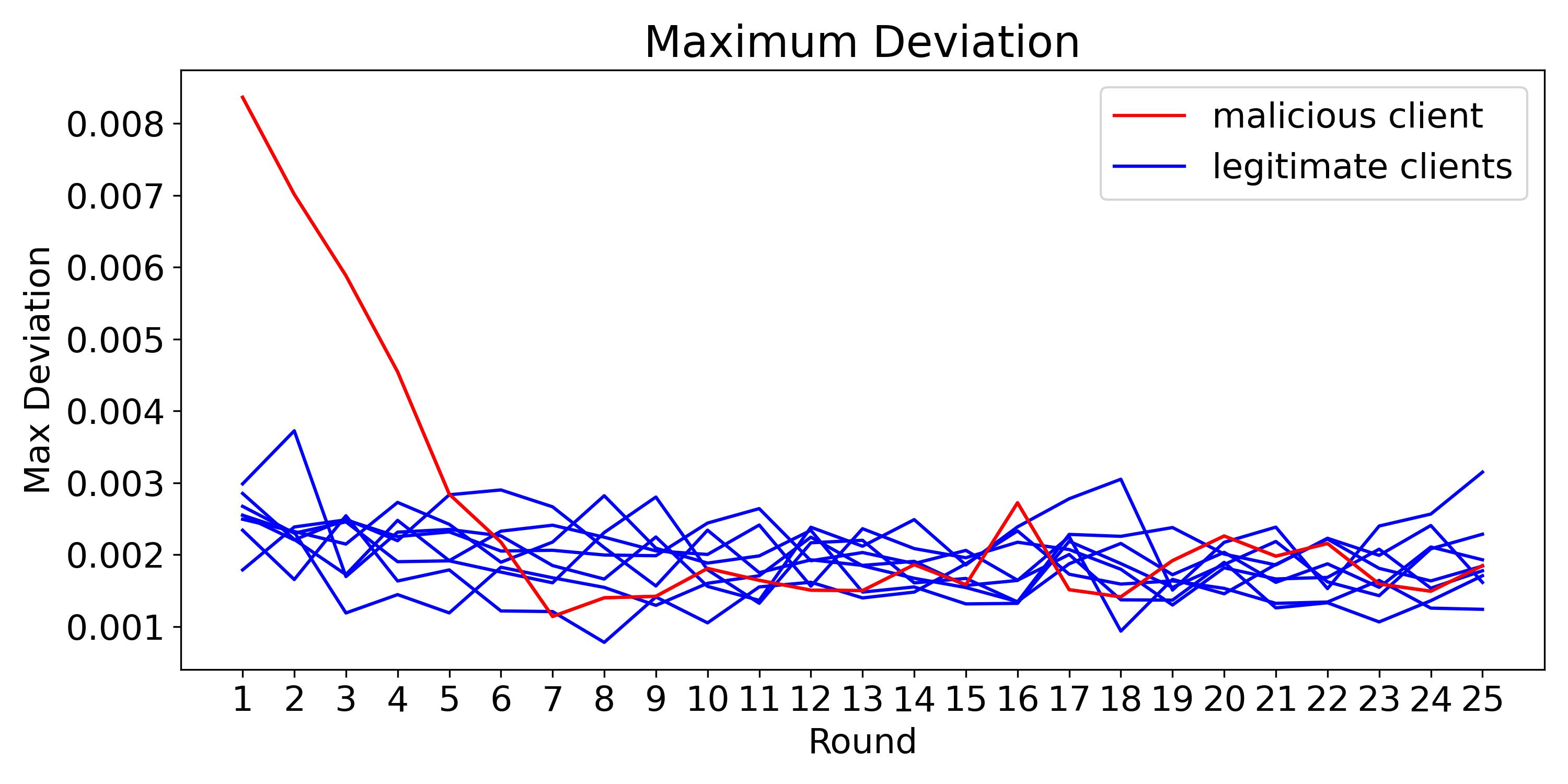}
    \end{subfigure}
    \begin{subfigure}[t]{0.32\textwidth}
        \includegraphics[width=\textwidth]{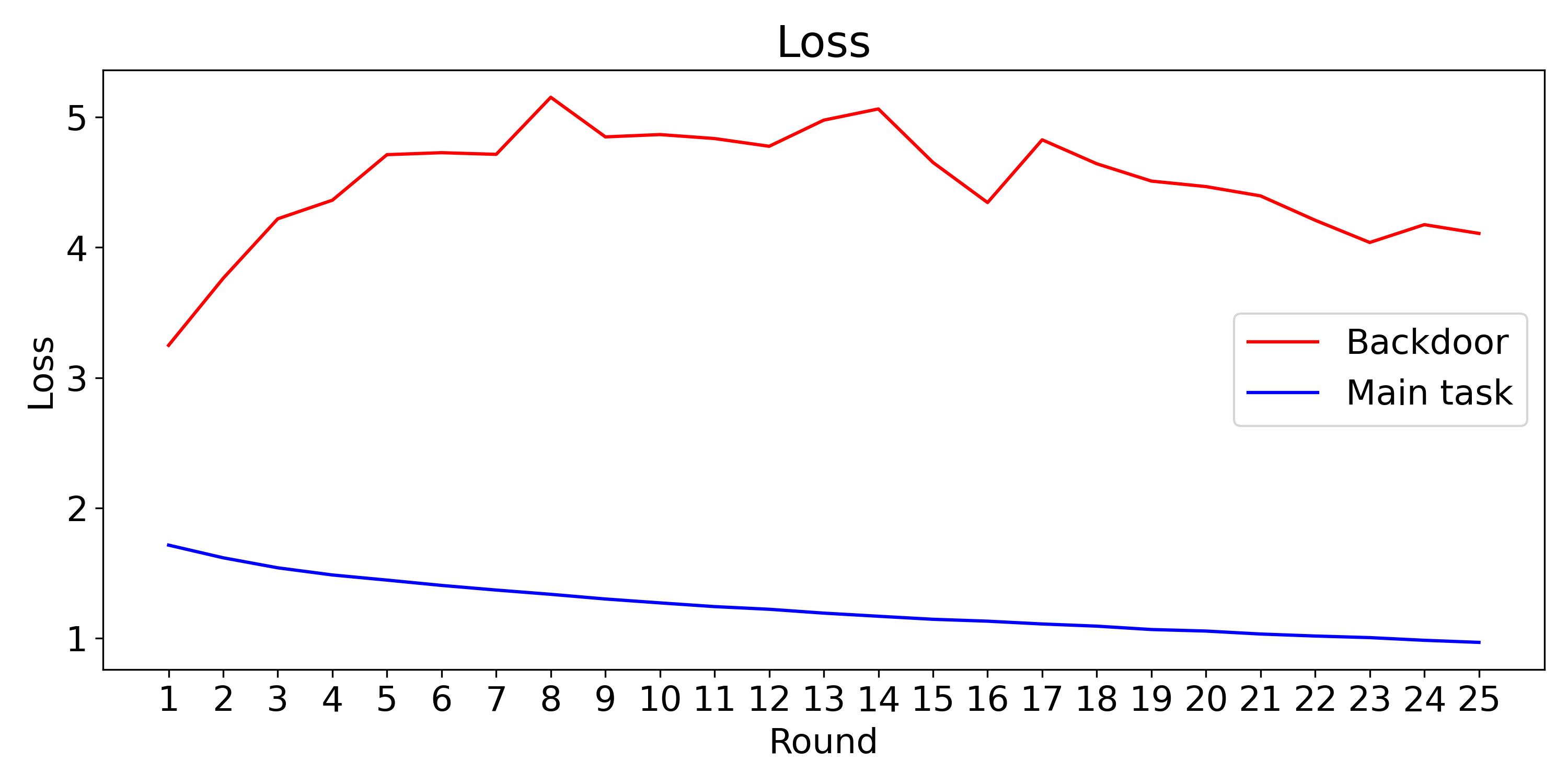}
        \caption{Malicious data rate = 1/3}
        \label{fig:mdr_b}
    \end{subfigure}
    \begin{subfigure}[t]{0.32\textwidth}
        \includegraphics[width=\textwidth]{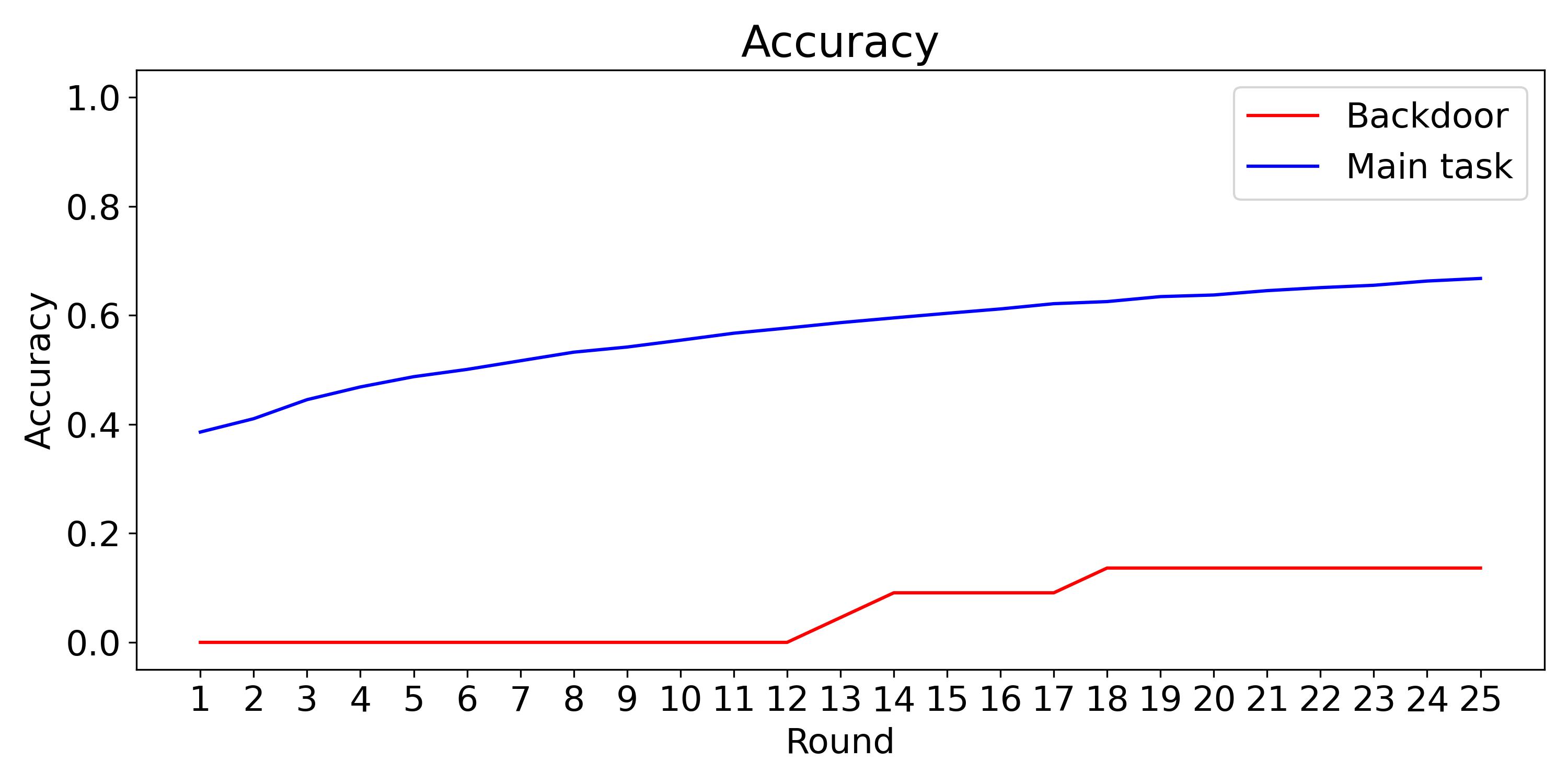}
    \end{subfigure}
    \begin{subfigure}[t]{0.32\textwidth}
        \includegraphics[width=\textwidth]{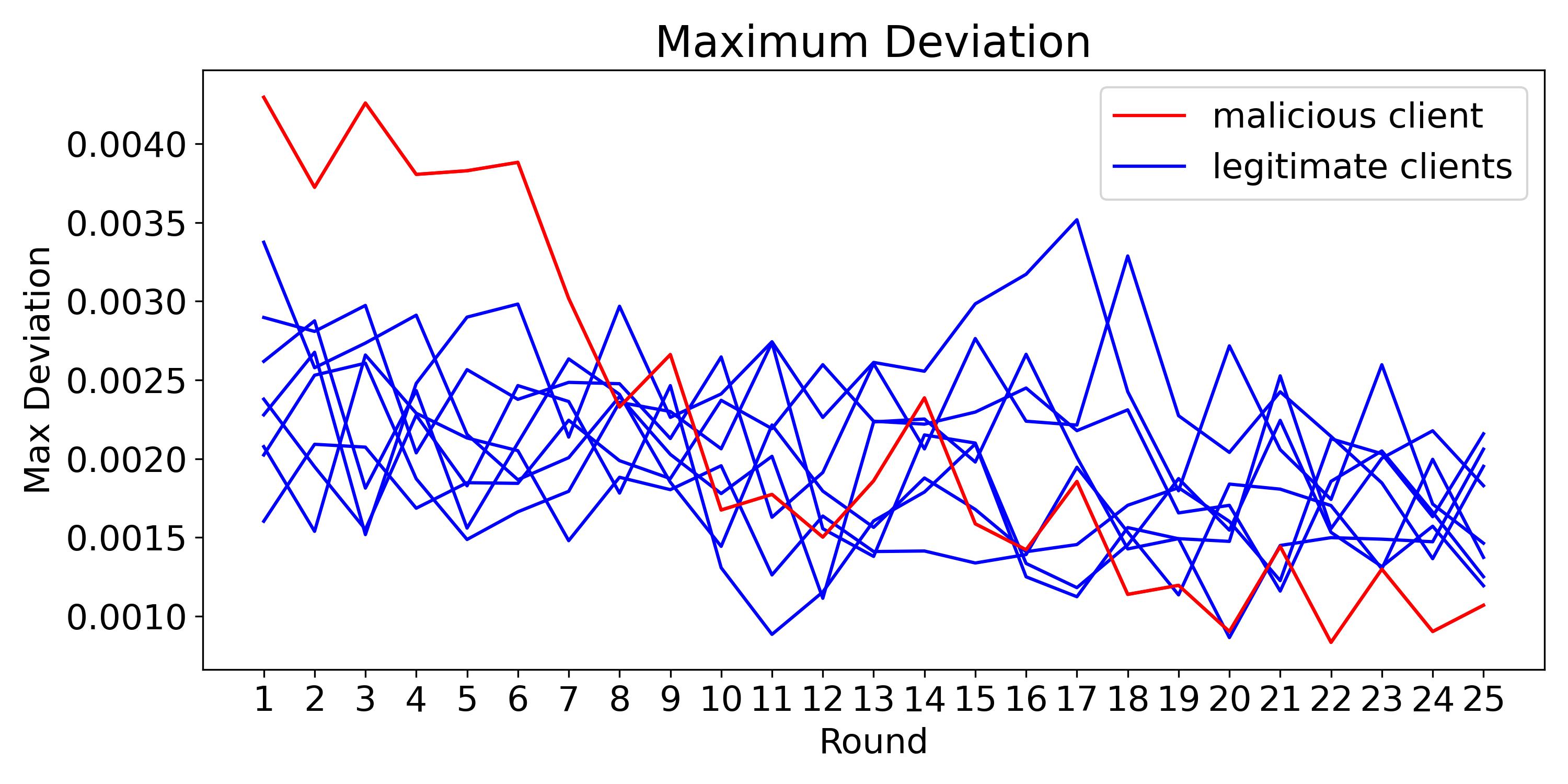}
    \end{subfigure}
    \begin{subfigure}[t]{0.32\textwidth}
        \includegraphics[width=\textwidth]{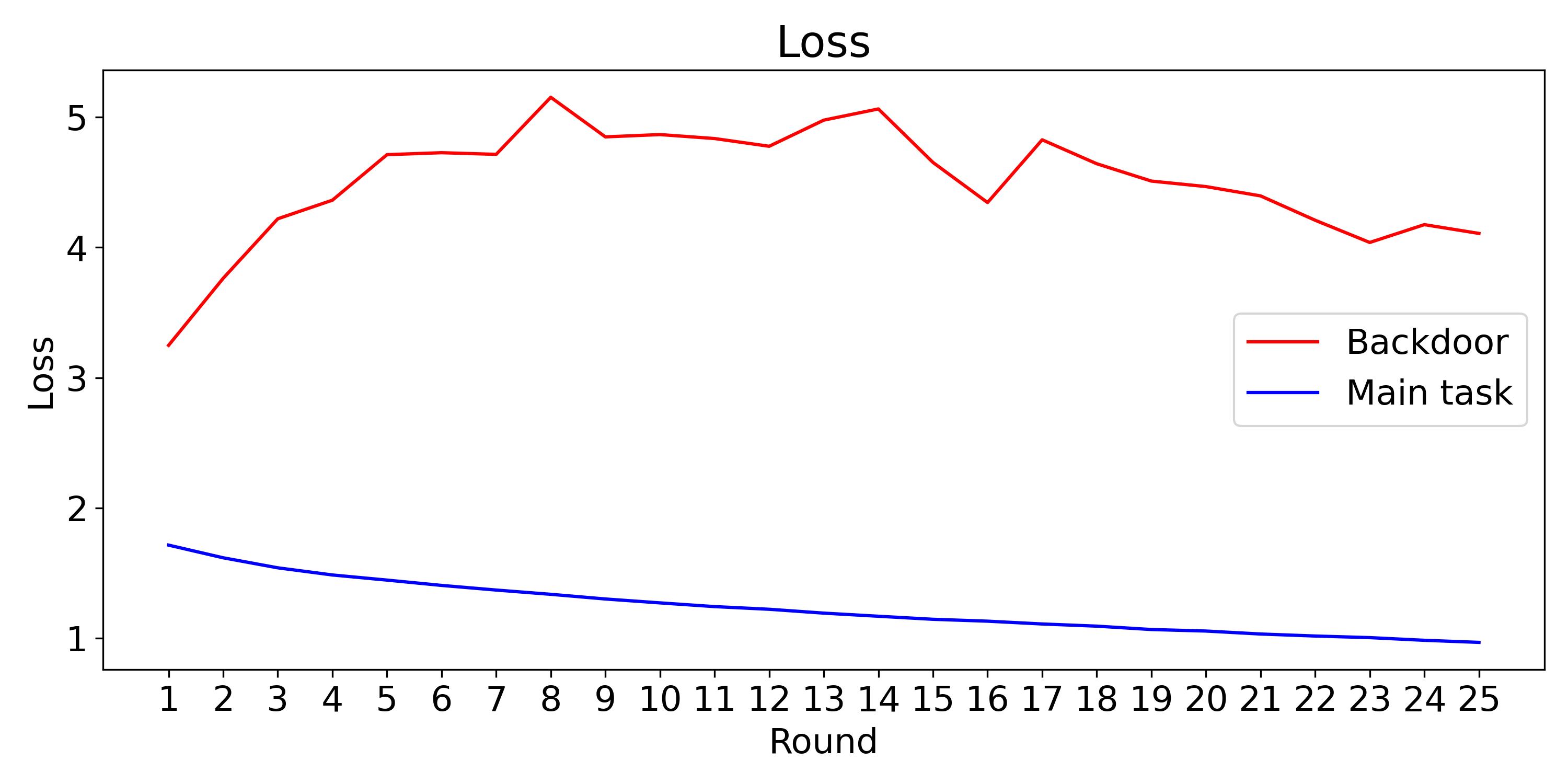}
        \caption{Malicious data rate = 1/10}
        \label{fig:mdr_c}
    \end{subfigure}
    \begin{subfigure}[t]{0.32\textwidth}
        \includegraphics[width=\textwidth]{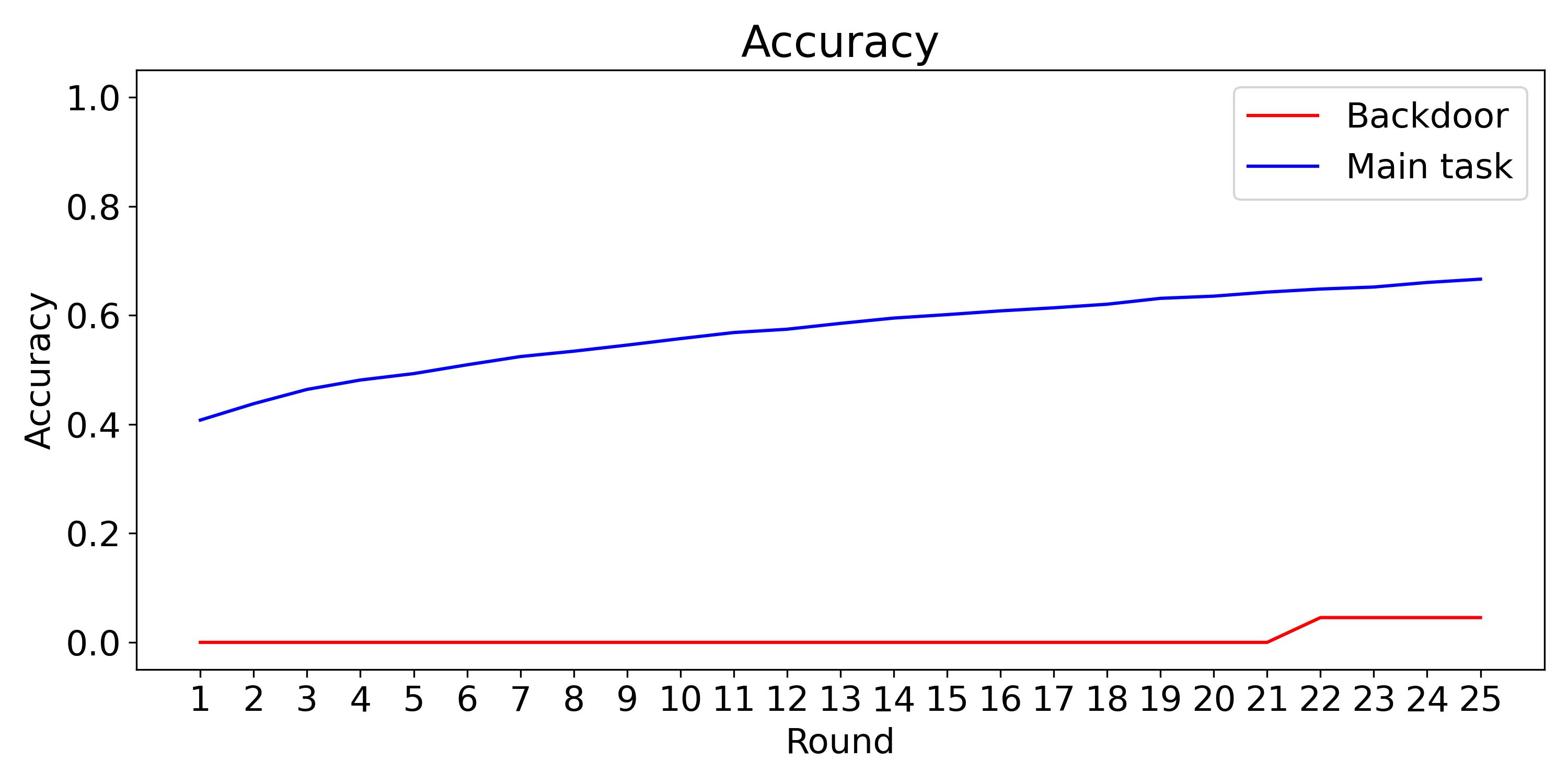}
    \end{subfigure}
    \caption{Impact of different malicious data rates of (a) 1/1, (b) 1/10, and (c) 1/50 on the maximum deviation in the last layer bias values, joint model loss and accuracy for both main task and backdoor in consecutive training rounds.}
    \label{fig:maliciousdatarate}
\end{figure*}

In this set of experiments, we investigate the impact of different malicious data rates on  deviations in local model weights. For this purpose, we adjust three different malicious data rates of 1/1, 1/10, and 1/50, which represent the ratio of malicious data to all data in malicious clients. We include one malicious client and set the learning rate to 0.01.

Figure \ref{fig:mdr_a} shows our experimental results for the malicious data rate of 1/1 which means all data is malicious in the malicious client (i.e. no legitimate data at all). According to Figure \ref{fig:mdr_a}, the maximum deviation in the malicious client is abnormally higher than the values seen in the legitimate clients in all rounds. Contrary to the trend seen in other experimental results, the maximum deviation value of the malicious client here does not decrease gradually but remains almost constant. The reason for this lies in the fact that gradients in backpropagation are calculated  as an average value over all samples in a batch, and therefore this gradient value remains almost constant given all samples are malicious and backdoor accuracy stays around the same level in all rounds (the opposite case is explained in Section \ref{sec:VaryingNumberMaliciousClients}). We can also infer from Figure \ref{fig:mdr_a} that even having 100\% malicious data is insufficient to successfully insert a backdoor into the joint model for the one-malicious client case. 

Figure \ref{fig:mdr_b} includes our experimental results for the malicious data rate of 1/10. It is seen from Figure \ref{fig:mdr_b} that maximum deviation in the malicious client's model takes abnormal values in the initial training rounds, and then gradually decreases as training proceeds and reaches the legitimate models' level at around the 6th training round. This is due to the convergence of local models and the joint model as well as due to legitimate samples in the malicious client.       

Figure \ref{fig:mdr_c} shows  our experimental results for the malicious data rate of 1/50. Notice the scale of the maximum deviation graph, which is much smaller than those in the other malicious data rate results,  which indicates that the magnitude of abnormal behavior of malicious client gets softened as malicious data rate decreases.

\section{Conclusion}
In this work, we study the problem of anomaly localization in model gradients under backdoor attacks against federated learning. To deepen our knowledge and understanding about the possible reflections of backdoor attacks on the low-level structure of neural networks models in various federated learning environments, we conduct both theoretical and experimental analysis to find, justify and validate exactly at which weights anomaly can be expected. In this sense, we discovered that it is more likely to observe backdoor-related anomalies in final layer bias weights of malicious local models. We also examined the possible impact of the number of malicious clients, learning rate, and malicious data rate on anomaly level.   Our findings can be used to optimize backdoor detecting methods such as making them focused on only the weights where the anomaly is most expected.

\bibliographystyle{unsrt}
\bibliography{references}

\end{document}